\documentclass[runningheads]{llncs}

\usepackage{eccv}

\usepackage{eccvabbrv}

\usepackage{graphicx}
\usepackage{booktabs}
\usepackage{amsmath,amsfonts,bm}       
\usepackage{algorithm}
\usepackage{algpseudocode}
\usepackage{multirow}
\usepackage{pifont}
\newcommand{\cmark}{\ding{51}}%
\newcommand{\xmark}{\ding{55}}%
\usepackage{makecell}
\usepackage{rotating}
\usepackage{xcolor}
\usepackage{stfloats}
\usepackage{wrapfig}
\usepackage{setspace}

\usepackage[accsupp]{axessibility}  

\usepackage[pagebackref,breaklinks,colorlinks,citecolor=eccvblue]{hyperref}

\usepackage{orcidlink}

\newcommand{\myparagraph}[1]{\vspace{4pt}\noindent{\bf #1}}

\begin{document}

\title{Just Add \$100 More: Augmenting NeRF-based Pseudo-LiDAR Point Cloud for Resolving Class-imbalance Problem} 

\titlerunning{Pseudo Ground Truth Augmentation}

\author{Mincheol Chang \inst{1}$^{\dagger}$\thanks{Work done during the internship at NAVER LABS} \and
Siyeong Lee \inst{2}$^{\dagger}$ \and
Jinkyu Kim\inst{1} \and 
Namil Kim\inst{2}$^{\ddagger}$ \\ 
$^{\dagger}$Equal contribution, $^{\ddagger}$Corresponding author
}


\authorrunning{M.~Chang et al.}


\institute{Korea University \\ \email{\{m1ncheoree, jinkyukim\}@korea.ac.kr} \and NAVER LABS \\ \email{\{siyeong.lee, namil.kim\}@naverlabs.com}}


\maketitle

\begin{abstract}
Typical LiDAR-based 3D object detection models are trained in a supervised manner with real-world data collection, which is often imbalanced over classes (or long-tailed). To deal with it, augmenting minority-class examples by sampling ground truth (GT) LiDAR points from a database and pasting them into a scene of interest is often used, but challenges still remain: inflexibility in locating GT samples and limited sample diversity. In this work, we propose to leverage pseudo-LiDAR point clouds generated (at a low cost) from videos capturing a surround view of miniatures or real-world objects of minor classes. Our method, called Pseudo Ground Truth Augmentation (PGT-Aug), consists of three main steps: (i) volumetric 3D instance reconstruction using a 2D-to-3D view synthesis model, (ii) object-level domain alignment with LiDAR intensity estimation and (iii) a hybrid context-aware placement method from ground and map information. 
We demonstrate the superiority and generality of our method through performance improvements in extensive experiments conducted on three popular benchmarks, i.e., \textbf{nuScenes}, \textbf{KITTI}, and \textbf{Lyft}, especially for the datasets with large domain gaps captured by different LiDAR configurations.
Our code and data will be publicly available upon publication.

\end{abstract}  
\section{Introduction}\label{sec:intro}
\begin{figure}[t]
    \centering
    \includegraphics[width=\linewidth]{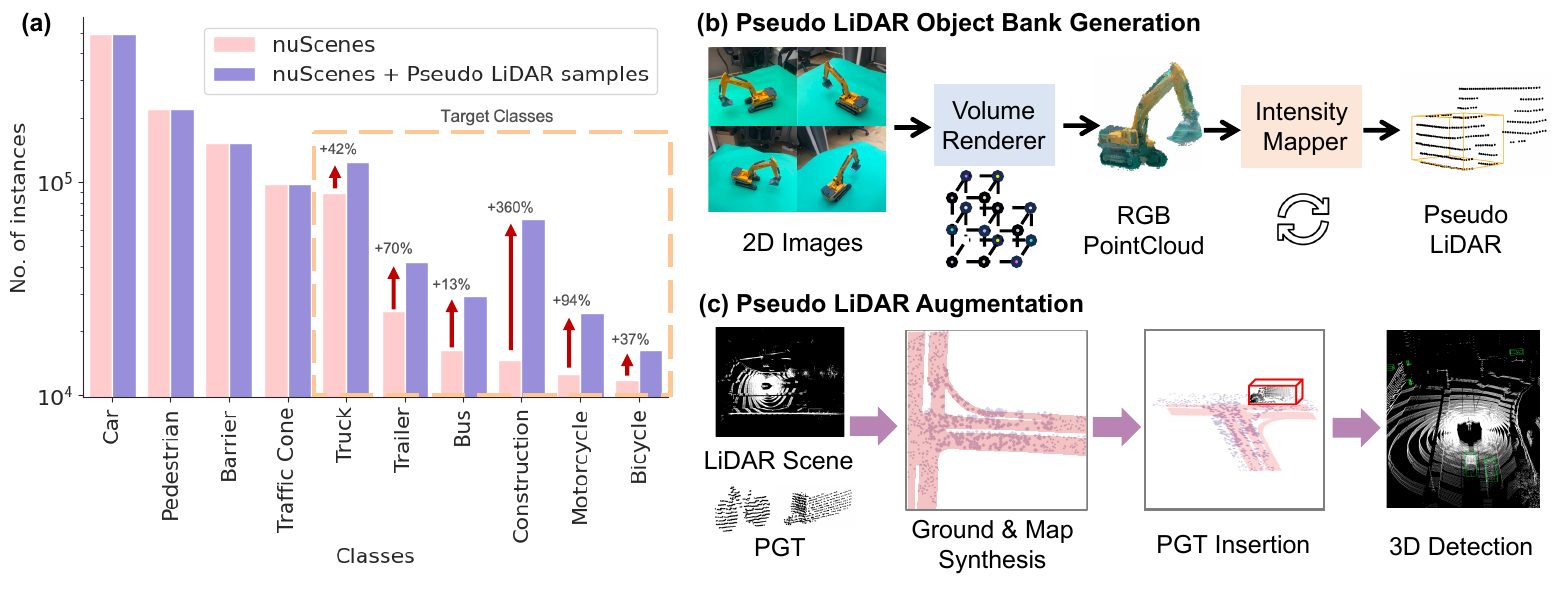}
    \caption{
    (a) Original nuScenes class distribution, which becomes more balanced by augmenting our generated pseudo-LiDAR samples for target minor classes. (b) Our proposed PGT-Aug reconstructs objects' 3D volumetric representation from multi-view images of miniatures and real-world objects of minor classes, followed by NeRF-based pseudo-LiDAR generation (see top). (c) These are then used to augment samples into a new scene to train LiDAR-based object detectors (see bottom).}
    \label{fig:stats}
\end{figure}

3D object detection has garnered growing interest thanks to its wide applications, such as autonomous driving and robotics.
LiDAR sensors are widely adopted in autonomous vehicles to measure 3D scene information as point clouds.
Many LiDAR-based 3D object detection models~\cite{DBLP:journals/sensors/YanML18, DBLP:conf/cvpr/ZhouT18, DBLP:conf/cvpr/LangVCZYB19, DBLP:conf/cvpr/ShiGJ0SWL20, DBLP:conf/cvpr/YinZK21, DBLP:conf/cvpr/ChenLZQJ23a, DBLP:conf/cvpr/WangSSLWHSW23} have been studied in terms of point cloud representations or model architectures.
However, there is a lack of attention to the class imbalance problem, including the reliability of rare objects in long-tail distributions.
This issue is not only prevalent in popular Autonomous Driving~(AD) datasets~(e.g., KITTI~\cite{DBLP:journals/ijrr/GeigerLSU13} and nuScenes~\cite{DBLP:conf/cvpr/CaesarBLVLXKPBB20}, see \cref{fig:stats} (a)) but a significant concern in the real world.

A naïve solution for the class imbalance problem is collecting more LiDAR data, but obtaining sufficient long-tail samples is practically challenging.
Due to the inherent imbalance in the natural distribution, the more data we collect, the imbalance between common and rare objects widens.
In existing methods~\cite{DBLP:journals/sensors/YanML18, DBLP:conf/cvpr/FangZZJWZ21, DBLP:journals/corr/abs-2309-05810}, the common strategy to deal with the class imbalance is a copy-and-paste-style sample augmentation, which copies target objects' points from the other frames within the dataset and pastes them into the current frame.
These objects have only been augmented with the original location.
To address such limitations, some researches~\cite{ DBLP:conf/cvpr/FangZZJWZ21, DBLP:journals/corr/abs-2305-12853} have been proposed to find the potential area where objects can be pasted.
However, the feasible positions are still restricted due to the lack of shape information at the acquisition time, and areas with low point cloud density remain unaddressed.

With advances in Neural Radiance Field (NeRF)~\cite{DBLP:journals/cacm/MildenhallSTBRN22}, recent studies~\cite{DBLP:conf/cvpr/Fridovich-KeilY22, DBLP:journals/tog/MullerESK22, DBLP:journals/tog/KerblKLD23} have shown that it is feasible to generate high-quality 3D rendered outputs from multiple viewpoints in low-cost computations.
NeRF-based frameworks can generate a 3D object fully visible in all 3D viewpoints; thus, We argue that the generated object can be more flexibly pasted anywhere in feasible positions of a given scene.
However, generating real-world 3D objects of minority classes with NeRF-based models may be challenging as we need images capturing an object from multiple viewpoints. 
Thus, we advocate leveraging miniatures and public videos to obtain 3D-rendered samples of target objects.
Note that miniatures are often used to achieve practical effects in the film industry, and collecting vehicle miniatures from various types could be practically feasible.

In this work, we introduce Pseudo Ground Truth Augmentation (PGT-Aug), which generates point clouds of minority-class objects from two sources: (i) surround-view videos of given miniatures and (ii) public videos of real-world objects. 
As shown in \cref{fig:stats} (b), we first utilize a 2D-to-3D renderer to reconstruct an object's 3D volumetric representation. 
Then, we transform it into LiDAR-like 3D point clouds, then rearrange/filter points and estimate their intensities, forming a pseudo LiDAR point clouds data collection. 
During training, such generated pseudo LiDAR points are sampled and placed into appropriate places of a scene without where-to-paste technical constraints. 
Our experiments with the public datasets demonstrate that our data augmentation with pseudo-LiDAR points effectively balances detection performance over minority classes.

Our contributions can be summarized into three-fold:
(1) We present PGT-Aug, a novel pseudo-LiDAR sample generation and augmentation pipeline for LiDAR-based object detection models, involving an effective yet low-cost solution to leverage multi-view images of miniatures and real-world objects. 
(2)	We present a novel domain alignment framework for view-agnostically generated Pseudo LiDAR objects. To reduce the discrepancy to the real samples, we simulated the spatial distribution via the filtering and rearrange technique and the LiDAR’s intensity value via the data-driven estimation model.
(3)	We present a novel map-aware composition that places pseudo LiDAR objects into appropriate locations based on the scene distribution. Our method can effectively operate across various scenes and outperform existing approaches.
\section{Related Work} \label{sec:related}

\myparagraph{LiDAR-based 3D Object Detection Datasets for Autonomous Driving.}
The AD community has seen a surge of interest in LiDAR-based 3D object detection due to its capability to overcome the limitations of camera-based object detectors for better system reliability.
However, creating a large-scale annotated dataset is challenging due to the high cost of sensors, spatial distribution differences depending on the sensor type, difficulty in annotation caused by reflection and distance variations, and object shape changes by occlusions.
Majorly, several datasets~\cite{DBLP:conf/cvpr/CaesarBLVLXKPBB20, DBLP:journals/corr/abs-2004-06320, DBLP:conf/corl/HoustonZBYCJOIO20,  DBLP:conf/nips/MaoNJLCLLY0LYXX21, DBLP:conf/eccv/MeiZYYQCK22} have been released by large-scale companies or institutions. 
Despite these efforts, the data quantity is still limited compared to the image domain, causing the lack of generalization and the class imbalance problem \cite{DBLP:journals/pami/OksuzCKA21, DBLP:journals/corr/abs-2303-00086}.
Instead of collecting large-scale datasets, we propose a novel approach of generating high-quality rare objects cheaply.
These generated objects can be used as data augmentation for existing detection pipelines, addressing the class imbalance problem.

\myparagraph{Data Augmentation of LiDAR Point Clouds.}
The difficulty for human 3D annotation has prompted many recent works to either focus on LiDAR simulation~\cite{DBLP:conf/corl/DosovitskiyRCLK17, DBLP:conf/cvpr/ManivasagamWWZS20, DBLP:conf/cvpr/FangZZJWZ21, DBLP:conf/eccv/ZyrianovZW22} which utilizes 3D rendering modules to simulate point cloud, or LiDAR augmentation method~\cite{DBLP:journals/sensors/YanML18, DBLP:journals/corr/abs-1908-09492, DBLP:journals/corr/abs-2305-12853} to scale up and better utilize existing point cloud.
Manivasagam et al.\cite{DBLP:conf/cvpr/ManivasagamWWZS20}, Fang et al. \cite{DBLP:journals/ral/FangZYZZMWY20}, and Fang et al. \cite{DBLP:conf/cvpr/FangZZJWZ21} attempt to reduce a domain gap between the simulated and real world by combining real-world data with synthetic objects or backgrounds when simulating LiDAR scans.
SHIFT3D~\cite{DBLP:journals/corr/abs-2309-05810} extended these methods by utilizing a deep SDF model to generate a point cloud of hard categories and refining the object's shape and pose adversarially.
R2DM~\cite{DBLP:journals/corr/abs-2309-09256} employ denoising diffusion probabilistic models to simulate whole input scenes using model-based approaches.
On the other hand, GT-Aug~\cite{DBLP:journals/sensors/YanML18} used a copy-and-paste strategy to insert the sample from other scenes into the current scene. Combining the realistic composition, Real-Aug~\cite{DBLP:journals/corr/abs-2305-12853} improved the data utilization and the performance. Instead of handling objects, PolarMix~\cite{DBLP:conf/nips/Xiao0GCL022} mixed scenes and instances from two different scenes to generate a new virtual scene. In this paper, we propose a novel real composition strategy and some techniques for data augmentation.

\myparagraph{Neural Radiance Fields and 3D Rendering.}
Many recent works attempt to represent 3D scenes as continuous implicit representations or differentiable structures~\cite{DBLP:conf/cvpr/ParkFSNL19, DBLP:conf/iccv/SaitoHNMLK19, DBLP:conf/eccv/PengNMP020, DBLP:journals/cacm/MildenhallSTBRN22, DBLP:conf/cvpr/Fridovich-KeilY22}. 
NeRF~\cite{DBLP:journals/cacm/MildenhallSTBRN22} represents 3D geometry by approximating density and view-dependent RGB using simple MLP architecture. 
Following NeRF, many works based on neural rendering have improved NeRF's speed and rendering quality~\cite{DBLP:conf/iccv/BarronMTHMS21, DBLP:conf/cvpr/BarronMVSH22, DBLP:journals/tog/MullerESK22, DBLP:conf/cvpr/Fridovich-KeilY22, DBLP:journals/tog/KerblKLD23}. 
Especially, Instant-NGP \cite{DBLP:journals/tog/MullerESK22} and Plenoxels~\cite{DBLP:conf/cvpr/Fridovich-KeilY22} contribute to faster training and rendering of 3D scenes.
Instant-NGP achieves this by leveraging multi-level hash tables for encodings to accelerate MLPs, while Plenoxels utilizes sparse voxel grids to interpolate view-dependent color and density fields.
With the notable progress in 3D differential rendering techniques, reconstructing 3D objects is now easier and more precise~\cite{DBLP:journals/tog/KerblKLD23}.
This allows these technologies to be effectively utilized for perception tasks \cite{DBLP:conf/nips/JeongSLCACP22}.
\begin{figure*}[t]
    \centering
    \includegraphics[width=0.95\linewidth]{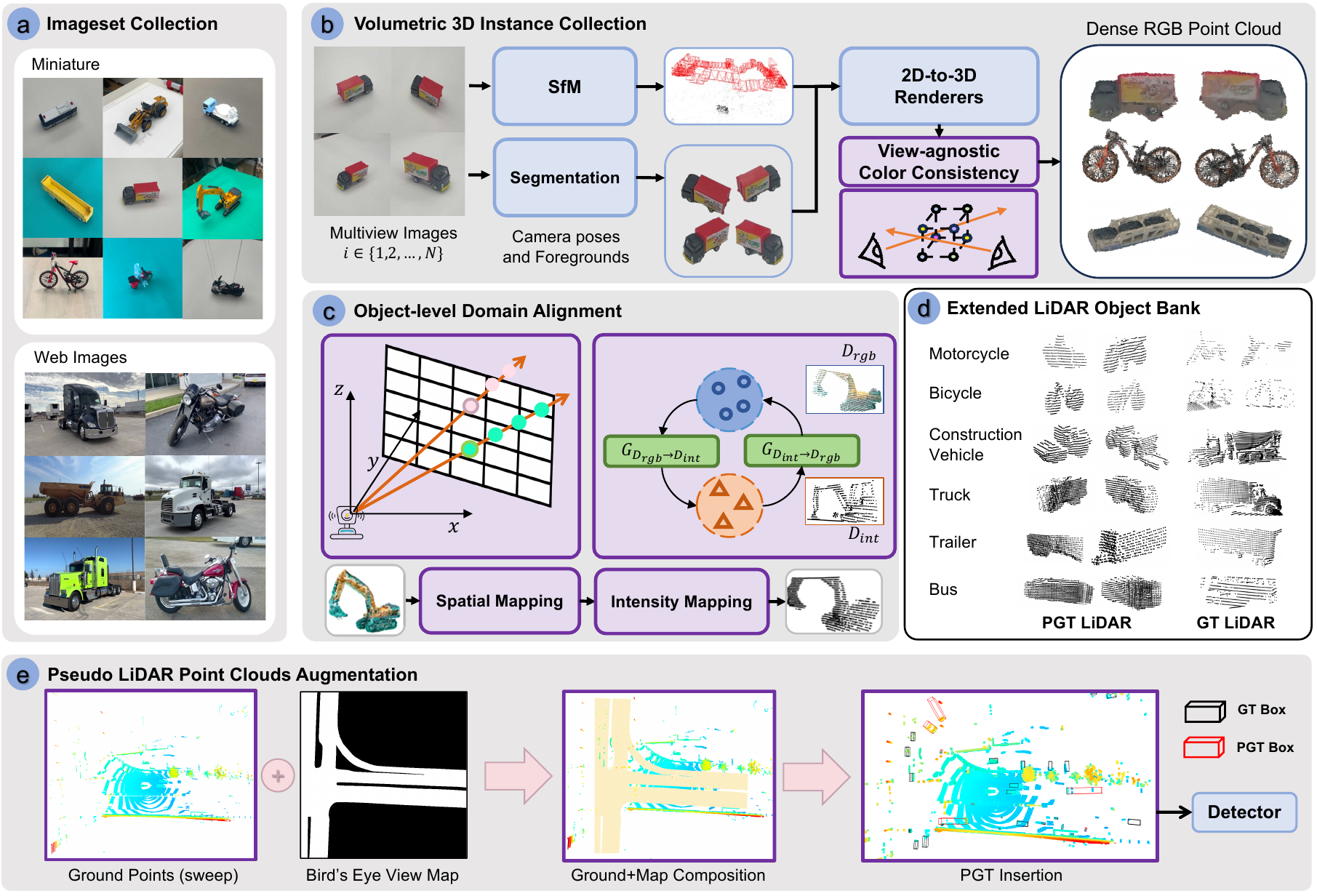}
    \vspace{-0.5em}
    \caption{\textbf{Overview of Pseudo GT (PGT)-Aug framework.} Given (a) surround-view images of miniatures or public videos of minority-class objects, (b) we first reconstruct their volumetric representations by estimating camera poses and foreground extraction, followed by 2D-to-3D rendering to obtain RGB-based objects' point clouds (\cref{sec:volumetric3d}). (c) We post-process RGB-based point clouds using spatial points rearrangement and CycleGAN-based intensity estimator (\cref{sec:viewdependent}), producing (d) view-dependent pseudo LiDAR point clouds. Such points are then stored in a bank with ground truth (GT) samples, and then (e) we paste sampled objects into the target scene based on hybrid information from the ground and map (\cref{sec:augmentation}). 
    }\label{fig:overview}
    \vspace{-1em}
\end{figure*}

\section{Method}\label{sec:method}
We introduce PGT-Aug, a fast, realistic, low-cost pipeline to render simulated LiDAR point clouds of a given real-world object. Such rendered 3D objects are then augmented to train LiDAR-based object detectors, reducing the class imbalance problem.
As shown in \cref{fig:overview}, our pipeline consists of three main modules: (i) \textbf{Volumetric 3D instance collection}, where we collect video frames of an arbitrary object and reconstruct its 3D volumetric representation (\cref{sec:volumetric3d}). (ii) \textbf{Object-level domain alignment}, where we transform the generated objects into realistic objects through sensor configurations and intensity estimation models. (\cref{sec:viewdependent}). Lastly, (iii) \textbf{Pseudo LiDAR Point Clouds Augmentation}, where we consider more realistic object insertion areas by using ground areas and map information and modify objects to align the dataset distribution (\cref{sec:augmentation}).

\subsection{Volumetric 3D Instance Collection}\label{sec:volumetric3d}
\myparagraph{Data Collection.} 
Traditional 3D object detection models, which often rely on human annotation and manual 3D modeling, struggled with scaling to out-of-domain objects.
To overcome this limitation, we present a study to augment desired objects using cost-effective image data, leveraging NeRF-based methods.
The easily obtainable data consists of \textbf{public videos}, but the availability of videos capturing $360^{\circ}$ surround view is limited. Therefore, we utilize readily obtainable \textbf{miniatures} that could be captured as shown in \cref{fig:overview} (a). As the title suggests, we purchased dozens of realistic miniatures \textit{for just \$100}. Consequently, the majority of our dataset utilizes miniatures, while the objects which were unavailable for purchase were generated from public videos.

\myparagraph{Preprocessing.}
As shown in \cref{fig:overview} (b), given collected video frames, we first estimate the following three information as a preprocessing for the later 2D-to-3D rendering: (i) a camera intrinsic matrix, (ii) camera poses (given by 3D camera position and its orientation for each frame) and (iii) binary masks for a foreground object.
For (i) and (ii), we use COLMAP~\cite{DBLP:conf/cvpr/SchonbergerF16}, similar to the preprocess of existing NeRF-based approaches~\cite{DBLP:journals/cacm/MildenhallSTBRN22, DBLP:journals/corr/abs-2010-07492, DBLP:conf/cvpr/Fridovich-KeilY22}.
For (iii), we use the off-the-shelf video segmentation model, i.e., Segment and Track Anything~\cite{DBLP:journals/corr/abs-2305-06558}, to extract segmentation masks for foreground objects in video frames.

\myparagraph{2D-to-3D Rendering.}
We build our framework upon the recent 2D-to-3D rendering methods.
For its high quality and efficiency in 3D reconstruction, we choose Plenoxels~\cite{DBLP:conf/cvpr/Fridovich-KeilY22} and Gaussian-Splatting~\cite{DBLP:journals/tog/KerblKLD23}, which are characterized by an explicit representation, differential pipeline, and no neural network-based method.
Therefore, the results of both models can easily converted to dense 3D point clouds and can be used to increase the number of out-of-distribution samples, unlike recent 3D conditional generative models, such as Shap-E~\cite{DBLP:journals/corr/abs-2305-02463} and Zero-1-to-3~\cite{DBLP:conf/iccv/LiuWHTZV23}, which are designed to represent the general traits in classes used for training.
Note that this part can be easily replaced with various rendering methods, and we mainly used Plenoxels in experiments to demonstrate baseline capability rather than generative capability.
\begin{figure}[t]
    \begin{center}
    \includegraphics[width=\linewidth]{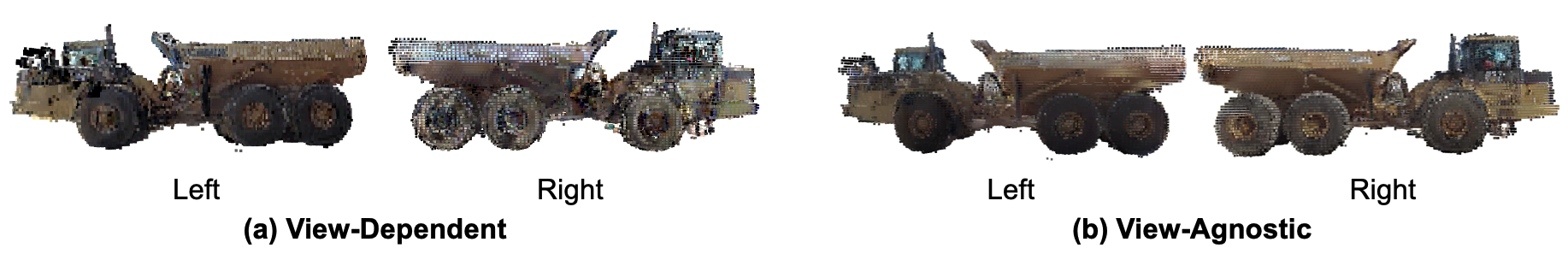}
    \end{center}
    \vspace{-2em}
    \caption{\textbf{Comparisons of View-Dependent and View-Agnostic Rendering.} 
    By maintaining a consistent number of colored points through view-agnostic rendering, we can create a view-agnostic bounding box for the given object.}
    \label{fig:viewagnostic}
\end{figure}

A series of Plenoxels~\cite{DBLP:conf/cvpr/Fridovich-KeilY22} and Gaussian-Splatting ~\cite{DBLP:journals/tog/KerblKLD23} methods originally predicts view-dependent representation. 
However, we need fully visible and uniformly high-density data observed from all view directions to create Pseudo LiDAR objects and axis-aligned bounding boxes.
Therefore, we propose an ad-hoc module to obtain representative colors for each voxel grid or point based on the estimated mean color values from different views.
More specifically, iterating through $N$ views of images, the module determines which voxel grid or point’s camera ray passes through and calculates the color value of voxels or points by doing a dot product between corresponding spherical harmonic coefficients and view-dependent harmonics basis.
If a certain voxel or a point is hit multiple times by rays from different views, we update the existing value by adding the new color value and store the total number of passes. 
After $N$ iterations, we derive the view-consistent color value by dividing the updated value by the stored value. 
This simple module facilitates the effective application of existing 2D-to-3D rendering methods for 3D object generation and can apply to a series of both Plenoxels~\cite{DBLP:conf/cvpr/Fridovich-KeilY22} and Gaussian-Splatting~\cite{DBLP:journals/tog/KerblKLD23} methods.

\myparagraph{Instance Size Determination.} We investigated the average size of objects for each class in the target dataset. We determine the size using Gaussian noise with $\sigma$ of 0.1. When the noise is out of one $\sigma$ range, it is clipped to the one $\sigma$.

\subsection{Object-level Domain Alignment} \label{sec:viewdependent}
Given the RGB-colored point clouds, our ultimate goal is to simulate them as minority-class samples for real LiDAR data augmentation. 
Unfortunately, our RGB-colored point clouds lack two types of information about the spatial distribution based on the sensor’s position, type and quantity and the power of a receiving signal~(Intensity), therefore are not fully descriptive for modeling and evaluation of LiDAR sensor with full properties.
Thus, we want to tackle this issue by formulating it as \textit{object-level domain alignment} between the collected RGB-colored point clouds $\mathcal{D}_{rgb}$ and the real-world point clouds $\mathcal{D}_{int}$, and propose two kinds of alignment techniques as (i) view-dependent points filtering and rearrangement and (ii) LiDAR intensity estimation.
\begin{figure}[t]
  \begin{minipage}[t]{.5\linewidth}
    \centering
    \includegraphics[width=1.0\linewidth]{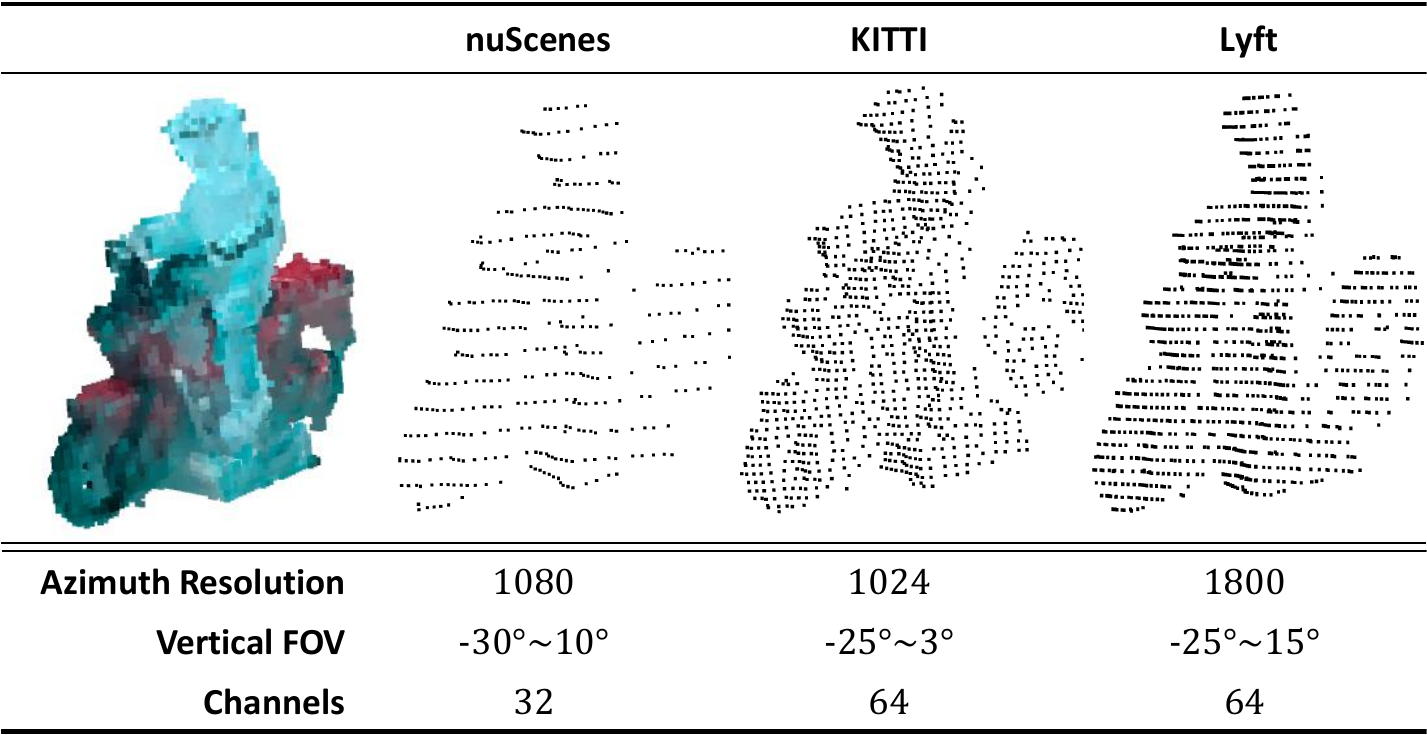}
    \captionof{figure}{\textbf{Effect of Rearranged Range Projection on Different Benchmarks.}}\label{fig:range_projection}
  \end{minipage}
  \hspace{1em}
  \begin{minipage}[t]{.45\linewidth}
    \centering
    \includegraphics[width=1.0\linewidth]{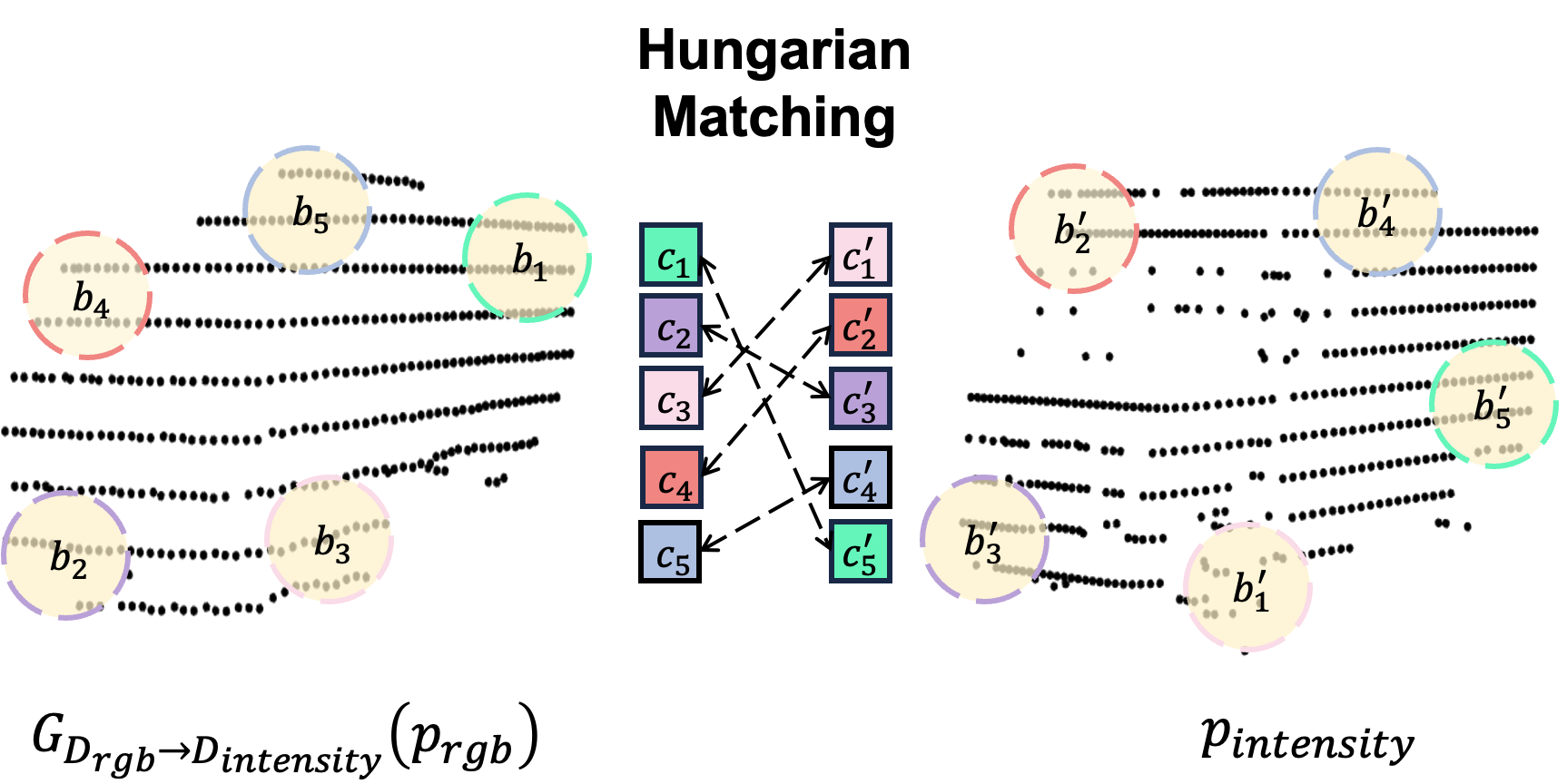}
    \captionof{figure}{\textbf{Illustrations of Ball Patch Matching.}}\label{fig:hungarian}
    \end{minipage}
\end{figure}

\myparagraph{View-dependent Points Filtering and Rearrangement.}
In this step, we aim to simulate realistic data variations based on the object’s position relative to LiDAR sensor location and configurations. Therefore, we require parameters for alignment modules based on factors such as type, position, quantity, and specifications~(Field of View and azimuth resolution) of LiDAR sensors in the target dataset.
We apply the following three steps: (i) \textbf{transformation} of points to the range view representation, (ii) \textbf{filtering} of points on the invisible side, and (iii) \textbf{reprojection} of points into 3D space.
In (i), given points in the spherical coordinates system, we map points into a range view $\mathbf{R}(u,v)\in \mathbb{R}^{H \times W}$, where $u$ and $v$ are the spatial grid indexes, i.e., grids of vertical and horizontal angle of points. 
For each grid $(u,v)$ in the range view, a point with minimum depth remains, and the others are filtered out (i.e., points on visible parts remain).
The remaining points are irregularly located in 3D space, generally different from those of real-world LiDAR point clouds.
We hence rearrange each point to be regularly spaced by adjusting their inclination $\phi$ and azimuth $\theta$ in the spherical coordinate system: $\phi' = \left[1 - \left(\frac{v+0.5}{H} \right)\right] \text{FOV}_{\text{total}} - \text{FOV}_{\text{down}}, \theta' = \pi \left[2\left(\frac{u+0.5}{W}\right) - 1\right]$, where $\text{FOV}_{\text{down}}$ and $\text{FOV}_{\text{total}}$ represent the down part and the total range of the field of view, respectively. Rearranged inclination and azimuth $\phi'$, $\theta'$ are calculated by back-projecting image coordinates $u, v$ to spherical coordinates following the above equations. Results on different LiDAR settings are illustrated on \cref{fig:range_projection}.

\myparagraph{CycleGAN-based LiDAR Intensity Estimation.}
Since the reflectivity of object surfaces cannot be extracted from the image data, the naïve way is designing the data-driven model for intensity estimation. As no real-world LiDAR points correspond to the generated RGB points, we develop an unpaired domain transfer model based on CycleGAN~\cite{DBLP:conf/iccv/ZhuPIE17}. Also, to directly reduce the intensity difference between two samples, we design a novel region matching loss that divides the objects into the same number of groups, even if the number of points varies between samples, as shown in~\cref{fig:hungarian}.
To compute the matching cost, both our generated point clouds~$G_{\mathcal{D}_{rgb}\to\mathcal{D}_{int}}(\mathbf{p}_{rgb})$ and real-world point clouds~$\mathbf{p}_{int}$ are grouped into an equal number of ball patches.
After dividing into $N$ ball patches, Hungarian matching is used to find the optimal assignment of ball patch pairs by computing the relative center distances for all matchings $\mathfrak{S}_N$,
\begin{equation}
    \hat{\sigma}=\underset{\sigma \in \mathfrak{S}_N}{\arg \min } \sum_i^N || c_{i} - c_{\sigma(i)}||_{1},
\end{equation}
where $N$ is the number of patches, $c_{i}$ and $c_{\sigma(i)}$ are centers of ball patches from $G_{\mathcal{D}_{rgb}\to\mathcal{D}_{int}}(\mathbf{p}_{rgb})$ and $\mathbf{p}_{int}$, respectively.
After finding optimal pairs, we reduce all intensity distances between fake and real pairs of patches.
\begin{equation}\label{eq:group_intensity}
    \mathcal{L}_{group} = \lambda \sum_j^N ||\mathbb{E}_{x \in b_{j}}(x) - \mathbb{E}_{y \in b_{\hat{\sigma}(j)}}(y)||_1,
\end{equation}
where $\lambda$ is a regularizing factor set to 0.1, and $\mathbb{E}_{x \in b_{j}}(x)$, $\mathbb{E}_{y \in b_{\hat{\sigma}(j)}}(y)$ denote the average of intensity values of optimal ball patch pairs $b_{j}, b_{\hat{\sigma}(j)}$.
The overall objective function is 
\begin{equation}
\mathcal{L}_{total} = \mathcal{L}_{\mathrm{CycleGAN}} + \mathcal{L}_{\mathrm{group}}.    
\end{equation}

\subsection{Pseudo LiDAR Point Clouds Augmentation}\label{sec:augmentation}
\begin{figure}[t]
    \centering
    \includegraphics[width=0.95\linewidth]{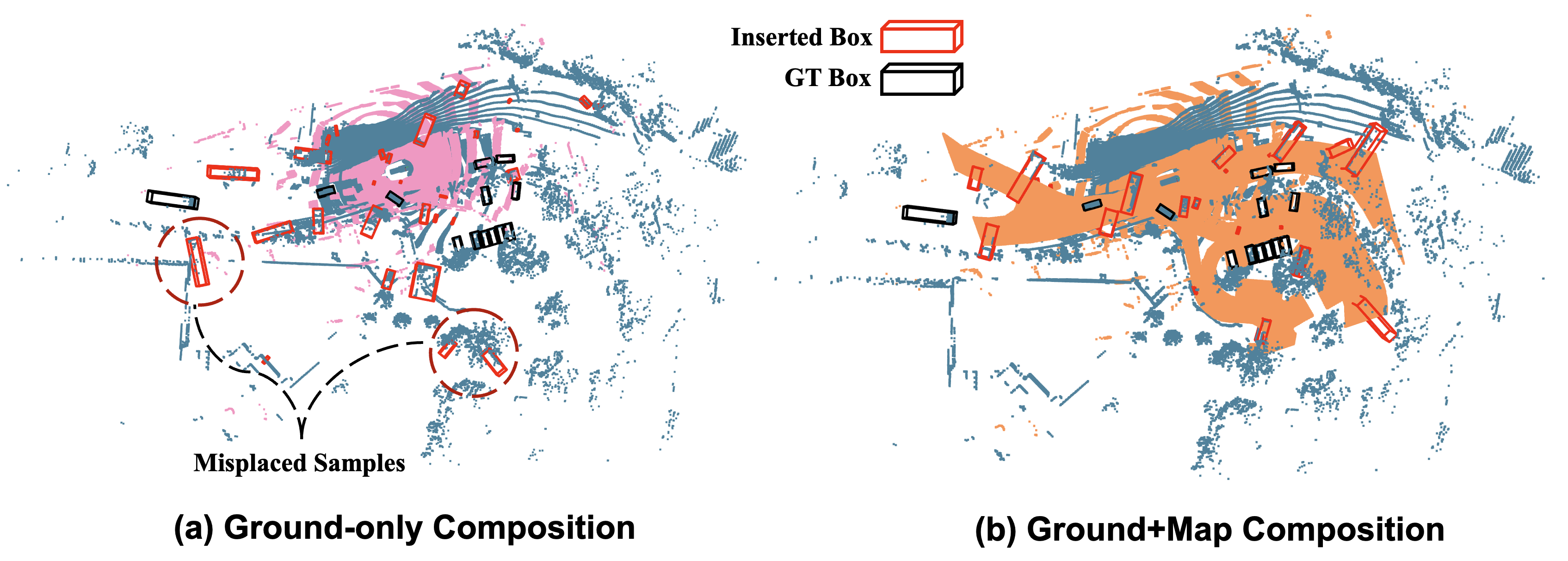}
    \vspace{-1em}
    \caption{\textbf{Comparison of Ground-only and Ground+Map Scene Composition.} (a) and (b) show ground-only and ground+map synthesized insertion results in a LiDAR scene. Pink and Orange-colored points denote the feasible location of insertion derived from (a) and (b) insertion methods respectively.} \label{fig:map}
    \vspace{-1em}
\end{figure}
\myparagraph{Ground and Map Synthesis for Object Insertion.}
To determine the feasible insertion areas, Real-Aug~\cite{DBLP:journals/corr/abs-2305-12853} and Lidar-Aug~\cite{DBLP:conf/cvpr/FangZZJWZ21} used estimated ground information.
As shown in~\cref{fig:map}, the estimated ground fails to cover insertable areas fully, so using map information is beneficial for realistic scene composition. 
However, since objects are not marked on the map, and not all objects are annotated in the dataset, we propose a simple yet effective placement area estimation method that utilizes map information and estimated ground areas.
We first obtain the map layout (e.g., Road, Sidewalk, etc.) within a radius of 51.2m around the ego vehicle and create a rasterized map with a 0.128m per pixel resolution.
Then, we select the proper layouts for the inserted object and fill the pixels in this area with 1. 
We also construct an additional rasterized ground map of the same size, assigning a pixel value of 1 in areas where the estimated ground points are located. 
The two pieces of information can potentially collide at the same pixel where dynamic objects are present, or the ground area is not estimated due to insufficient points.
Therefore, we use the map's value for pixels where the density of LiDAR points is low; otherwise, we use the estimated ground's value.
As shown in~\cref{fig:map}-(b), the proposed method can predict broader and feasible areas for more realistic data augmentation than ground-only composition.

\myparagraph{Aligning to Data Geometry.}
A generated object must be aligned in the axis for annotating the object with accurate bounding boxes in the detection downstream task.
The generated objects must be aligned on the same axis as the real annotated objects for pseudo labels in detection tasks.
For this, we first use PCA to align the generated points and rotate them around the normal direction in $xy$, $yz$, and $zx$ planes to align its axes.
We also need to classify objects’ front and back for heading information, so we adopt PointNet++~\cite{DBLP:conf/nips/QiYSG17} as a heading classifier trained using binary cross-entropy loss.
Since PointNet++ is rotation sensitive \cite{DBLP:journals/pr/ZhaoYXZCL22} and is trained without rotation augmentation during the training, the model can distinguish between front and back.

\myparagraph{Virtual Object Sweeps.} 
Sweeping LiDAR scans is a common way to increase the amount of point density in some popular benchmarks~\cite{DBLP:journals/ijrr/GeigerLSU13, DBLP:conf/cvpr/CaesarBLVLXKPBB20}.
Since the pipeline can generate objects that appear from a single scan, additional techniques are required to apply them to such datasets.
We introduce a rigid body motion model to create stacked object points along the time axis.
Given the dataset, we first collect the velocity and acceleration of each class's center points at each time step and estimate a motion trajectory.
With a set of trajectories, we virtually translate the center of generated object points along the selected trajectory. 
Ultimately, we can generate a virtual point cloud set that closely resembles what is acquired from the real LiDAR sweeps. 
\section{Experiments}

\myparagraph{Implementation Details.}
We use class-aware PointNeXt~\cite{DBLP:conf/nips/QianLPMHEG22} architecture for the generator and discriminator in our CycleGAN-based LiDAR intensity estimation module. 
To train such a module, we sample 600 (each for 300) object-wise point sets from our generated pseudo-LiDAR and nuScenes samples using the farthest point sampling approach.
Note that we filter out samples with less than 256 points.
Also, we use implementations of LiDAR-based 3D object detection from OpenPCDet~\cite{openpcdet2020} without any parameter changes.
All detectors are trained with a batch size of 32 on 4$\times$A100 GPUs during 20 epochs.
For a fair comparison, we conducted all experiments on the fixed seed.
The supplemental material provides more implementation details (e.g., architectures and hyperparameters).

\begin{figure*}[t]
    \begin{center}
    \includegraphics[width=0.95\linewidth]{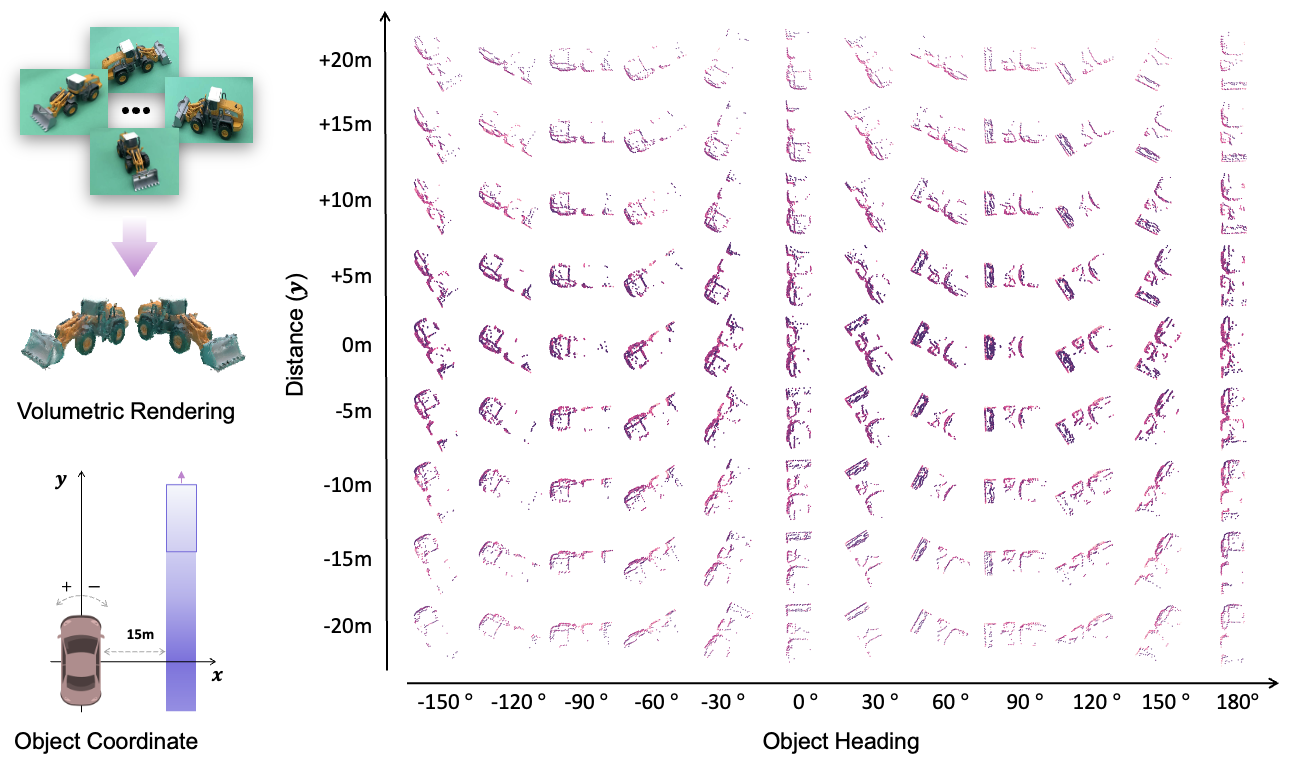}
    \end{center}
    \vspace{-1em}
    \caption{\textbf{Examples of generated pseudo-LiDAR point samples} with different orientations and ranges given reconstructed 3D volumetric representations.}\label{fig:example-of-pseudo-lidar}
    \vspace{-1em}
\end{figure*}
\myparagraph{Pseudo Object Bank Details.}
To validate our system, we have chosen nuScenes~\cite{DBLP:conf/cvpr/CaesarBLVLXKPBB20} dataset.
For the pseudo object bank, we generated samples to insert objects in 13 heading directions~(from $-180^{\circ}$ to $180^{\circ}$ at intervals of $30^{\circ}$) within the entire perception range~(from -50m to 50m at intervals of 5 meters).
All samples have their intensities through our intensity estimator. 
During training, we discard objects with less than 16 points to prevent ambiguity among classes and add a small variation in the center and size components to increase the diversity of the sample. 
As a result, our bank has 36,960 trucks, 52,800 construction vehicles, 12,960 buses, 19,280 trailers, 25,279 motorcycles, and 4,300 bicycles.
\begin{table}[t]
    \centering
    \caption{\textbf{Detection performance comparison on nuScenes \textit{val} set in terms of AP, mAP, and NDS.} Based on CV-Voxel~\cite{DBLP:conf/cvpr/YinZK21}, we compare different placement methods, such as random~\cite{DBLP:journals/sensors/YanML18}, ground-based~\cite{DBLP:journals/corr/abs-2305-12853}, and our map-based placement. We also report the effect of using our generated pseudo-LiDAR samples.}\label{tab:nerf_effect}
    \vspace{-1em}
    \resizebox{\linewidth}{!}{
    \begin{tabular}{lc|cc|cc|cc}
    \toprule
    \multicolumn{2}{l|}{GT-Aug~\cite{DBLP:journals/sensors/YanML18} (Random Placement)} & \cmark & \cmark & \xmark & \xmark & \xmark & \xmark\\
    \multicolumn{2}{l|}{Real-Aug~\cite{DBLP:journals/corr/abs-2305-12853} (Ground-based Placement)} & \xmark & \xmark & \cmark & \cmark & \xmark & \xmark \\
    \multicolumn{2}{l|}{Ours (Ground+Map-based Placement) } & \xmark & \xmark & \xmark & \xmark & \cmark & \cmark \\\midrule
    \multicolumn{2}{l|}{w/ Pseudo-LiDAR Samples} & \xmark & \cmark & \xmark & \cmark & \xmark & \cmark\\
    \midrule \midrule
    Class & \# of objects in \textit{val} & \multicolumn{6}{c}{Performance} \\
    \midrule
    Bus & 3009 & 69.1 & 71.3 \scriptsize{(+2.2\%)} & 71.8 & 71.7 \scriptsize{(-0.1\%)} & \textbf{73.0} & 72.1 \scriptsize{(-0.9\%)} \\
    Construction Vehicle & 2387 & 18.0 & 18.1 \scriptsize{(+0.1\%)} & 22.7 & 24.0 \scriptsize{(+1.3\%)} & 23.7 & \textbf{24.2} \scriptsize{(+0.5\%)} \\
    Trailer & 3765 & 36.1 & 35.7 \scriptsize{(-0.4\%)} & 41.6 & 41.4 \scriptsize{(-0.2\%)} & \textbf{42.3} & 40.4 \scriptsize{(-1.9\%)} \\
    Truck & 13950 & 56.9 & 57.8 \scriptsize{(+0.9\%)} & 57.3 & 58.3 \scriptsize{(+1.0\%)} & 58.8 & \textbf{59.8} \scriptsize{(+1.0\%)} \\
    Motorcycle & 2227 & 60.3 & 61.5 \scriptsize{(+1.2\%)} & 67.1 & 68.0 \scriptsize{(+0.9\%)} & 68.7 & \textbf{70.3} \scriptsize{(+1.6\%)} \\
    Bicycle & 2071 & 41.5 & 45.1 \scriptsize{(+3.6\%)} & 54.9 & 57.1 \scriptsize{(+2.2\%)} & 56.7 & \textbf{58.3} \scriptsize{(+1.6\%)} \\
    \midrule
    \multicolumn{2}{c|}{mAP ($\uparrow$, \textit{for all 10 classes})} & 59.0 & 59.8 \scriptsize{(+0.8\%)} & 62.3 & 63.0 \scriptsize{(+0.7\%)} & 63.3 &\textbf{63.5} \scriptsize{(+0.2\%)} \\
    \multicolumn{2}{c|}{NDS ($\uparrow$, \textit{for all 10 classes})}& 66.5 & 66.9 \scriptsize{(+0.4\%)} & 68.4 & 68.7 \scriptsize{(+0.3\%)} & 68.7 &\textbf{69.1} \scriptsize{(+0.4\%)} \\ 
    \bottomrule
    \end{tabular}}
\end{table}

\myparagraph{NuScenes \textit{val} set.} 
We report the effect of leveraging pseudo-LiDAR point clouds as GT samples for data augmentation, i.e., a sample from a collection of pseudo-LiDAR points is placed onto a scene during a training phase.
For this, we conduct all experiments here based on CP-Voxel~\cite{DBLP:conf/cvpr/YinZK21} with a voxel size of $[0.075, 0.075, 0.2]$.
We observe in \cref{tab:nerf_effect} that augmenting pseudo-LiDAR points into the GT database clearly improves the overall detection performance (including minor classes) regarding mAP and NDS (compare 1st vs. 2nd, 3rd vs. 4th, and 5th vs. 6th columns).
Moreover, our placement approach outperforms existing approaches, such as random and ground-based placement (compare 1st, 3rd vs. 5th columns).
Combining our two methods achieves the best performance, 69.1\% in NDS and 63.5\% in mAP.
This may confirm that our high-quality pseudo-LiDAR samples help complement the detection performance of minority classes without degradation in majority classes.
\begin{table}[!ht]
    \centering
    \caption{\textbf{Comparisons between baseline and PGT-Aug for individual models on nuScenes \textit{val} set.} \textit{Abbr.} C.V: Construction Vehicle, Ped: Pedestrian, T.C: Traffic Cone, M.C: Motorcycle, B.C: Bicycle. $^\dagger$: our reproduction.} \label{tab:model_robustness} 
    \vspace{-1em}
    \resizebox{\linewidth}{!}{
    \begin{tabular}{ll|cccccccccc|c|c}
    \toprule
    \multirow{3}{*}{Model} & \multirow{3}{*}{Aug.} & \multicolumn{4}{c|}{Majority classes} & \multicolumn{6}{c|}{Minority classes} & \multirow{3}{*}{mAP} & \multirow{3}{*}{NDS}\\
    \cmidrule{3-12}
     &  & Car & Ped & Barrier & \multicolumn{1}{c|}{T.C} & Bus & C.V & Trailer & Truck & M.C & \multicolumn{1}{c|}{B.C} & \\
    \midrule \midrule
    \multirow{3}{*}{SECOND~\cite{DBLP:journals/sensors/YanML18}} & GT-Aug & 81.5 & 77.3 & 58.3 & \multicolumn{1}{c|}{57.9} & 67.2 & 15.2 & 36.2 & 50.6 & 40.7 & \multicolumn{1}{c|}{16.2} & 50.11 & 61.56 \\
      & Real-Aug$^\dagger$ & 84.5 & \textbf{80.1} & 61.8 & \multicolumn{1}{c|}{\textbf{67.4}} & \textbf{72.0} & 24.1 & \textbf{44.2} & 58.7 & 61.5 & \multicolumn{1}{c|}{36.6} & 59.09 & 67.23 \\
      & PGT-Aug & \textbf{84.8} & 80.0 & \textbf{62.2} & \multicolumn{1}{c|}{67.2} & 71.9 & \textbf{24.4} & 42.4 & \textbf{58.8} & \textbf{64.0} & \multicolumn{1}{c|}{\textbf{38.1}} & \textbf{59.37} & \textbf{67.30} \\
    \midrule
    \multirow{3}{*}{CP-Pillar~\cite{DBLP:conf/cvpr/YinZK21}} & GT-Aug & 83.1 & 82.5 & 65.0 & \multicolumn{1}{c|}{65.8} & 63.0 & 14.1 & 23.7 & 54.4 & 51.4 & \multicolumn{1}{c|}{25.5} & 52.85 & 62.57 \\
      & Real-Aug$^\dagger$ & 82.9 & \textbf{84.0} & 65.7 & \multicolumn{1}{c|}{\textbf{67.9}} &  64.9 & 19.5 & 26.3 & \textbf{57.2} & 64.6 & \multicolumn{1}{c|}{46.2} & 58.03 & 64.85 \\
      & PGT-Aug & \textbf{83.3} & 83.5 & \textbf{65.7} & \multicolumn{1}{c|}{67.4} & \textbf{66.0} & \textbf{22.1} & \textbf{27.5} & 56.3 & \textbf{65.0} & \multicolumn{1}{c|}{\textbf{47.0}} & \textbf{58.39} & \textbf{65.49} \\
    \midrule
    \multirow{3}{*}{CP-Voxel~\cite{DBLP:conf/cvpr/YinZK21}} & GT-Aug & 84.9 & \textbf{85.4} & \textbf{68.3} & \multicolumn{1}{c|}{69.9} & 69.1 & 18.0 & 36.1 & 56.9 & 60.3 & \multicolumn{1}{c|}{41.5} & 59.04 & 66.54 \\
      & Real-Aug$^\dagger$ & 84.7 & 85.0& 67.5 & \multicolumn{1}{c|}{70.0} & 71.8 & 22.7 & \textbf{41.6} & 57.3 & 67.1 & \multicolumn{1}{c|}{54.9} & 62.27 & 68.39 \\
      & PGT-Aug & \textbf{85.4} & \textbf{85.4} & 68.0 & \multicolumn{1}{c|}{\textbf{71.1}} & \textbf{72.1} & \textbf{24.2} & 40.4 & \textbf{59.8} & \textbf{70.3} & \multicolumn{1}{c|}{\textbf{58.3}} & \textbf{63.52} & \textbf{69.11} \\
    \midrule
    \multirow{3}{*}{Transfusion-L~\cite{DBLP:conf/cvpr/BaiHZHCFT22}} & GT-Aug & 86.6 & 86.6 & 69.4 & \multicolumn{1}{c|}{73.6} & 72.7 & 23.3 & 43.9 & 53.3 & 69.4 & \multicolumn{1}{c|}{56.1} & 63.48 & 68.58 \\
     & Real-Aug$^\dagger$ & 86.6 & \textbf{86.9} & \textbf{69.9} & \multicolumn{1}{c|}{73.6} & 73.1 & 25.9 & 42.2 & \textbf{53.8} & 68.3 & \multicolumn{1}{c|}{55.1} & 63.54 & 68.57\\
     & PGT-Aug & \textbf{87.0} & 86.4 & 69.4 & \multicolumn{1}{c|}{\textbf{74.0}} &  \textbf{73.4} & \textbf{26.7} & \textbf{46.9} & 50.6 & \textbf{71.3} & \multicolumn{1}{c|}{\textbf{56.3}} & \textbf{64.20} & \textbf{68.83}\\
    \midrule
    \multirow{3}{*}{VoxelNeXt~\cite{DBLP:conf/cvpr/ChenLZQJ23a}} & GT-Aug & 83.7 & 84.5 & \textbf{68.9} & \multicolumn{1}{c|}{68.4} &71.4  & 20.9 & 37.5 & 56.1 & 62.9 &  \multicolumn{1}{c|}{49.8} & 60.42 & 67.03 \\
      & Real-Aug$^\dagger$ & 83.4 & 85.0& 67.8 & \multicolumn{1}{c|}{69.6} & 70.9 & 22.6 & 38.7 & \textbf{58.0} & 69.8 & \multicolumn{1}{c|}{\textbf{56.7}} & 62.25 & 67.62\\
      & PGT-Aug & \textbf{84.2} & \textbf{85.0} & 67.7 & \multicolumn{1}{c|}{\textbf{70.6}} & \textbf{72.4} & \textbf{23.9} & \textbf{40.6} & 57.5 & \textbf{70.9} &  \multicolumn{1}{c|}{56.6} & \textbf{62.95} & \textbf{68.30} \\
    \bottomrule
    \end{tabular}}
\end{table}

\myparagraph{Model Architectures.} 
We also conduct experiments comparing detection performance between GT-Aug, Real-Aug, and PGT-Aug based on five different detection baselines: SECOND~\cite{DBLP:journals/sensors/YanML18}, CP-pillar~\cite{DBLP:conf/cvpr/YinZK21}, CP-voxel~\cite{DBLP:conf/cvpr/YinZK21}, Transfusion-L~\cite{DBLP:conf/cvpr/BaiHZHCFT22}, and VoxelNeXt~\cite{DBLP:conf/cvpr/ChenLZQJ23a}. 
As shown in \cref{tab:model_robustness}, PGT-Aug brings significant improvements over other augmentations in all types of models, such as center-based models (CP-Voxel~\cite{DBLP:conf/cvpr/YinZK21}, CP-Pillar~\cite{DBLP:conf/cvpr/YinZK21}), the anchor-based model (SECOND~\cite{DBLP:journals/sensors/YanML18}), and other types (VoxelNeXt~\cite{DBLP:conf/cvpr/ChenLZQJ23a}, Transfusion-L~\cite{DBLP:conf/cvpr/BaiHZHCFT22}). 
This confirms that our method can generally be applied to various baseline detection models, boosting the model's accuracy for minority classes without sacrificing accuracy for majority classes. 
Note that SECOND~\cite{DBLP:journals/sensors/YanML18} was performed on a voxel with a size of $[0.1, 0.1, 0.2]$. The CP-Pillar~\cite{DBLP:conf/cvpr/YinZK21} was performed on a voxel with a size of $[0.1, 0.1, 8]$, while the other remaining models were performed with a voxel size of $[0.075, 0.075, 0.2]$. 

{
\setlength{\tabcolsep}{3pt}
\renewcommand{\arraystretch}{1}
\begin{table}[t]
    \centering
    \caption{\textbf{Detection performance comparison on nuScenes \textit{test} set in terms of AP, mAP, and NDS}. We use CP-Voxel~\cite{DBLP:conf/cvpr/YinZK21} and Transfusion-L~\cite{DBLP:conf/cvpr/BaiHZHCFT22} as a baseline model. $^\dagger$: our reproduction, $^\ddagger$: test time augmentation enabled.}\label{tab:test}
    \vspace{-1em}
    \begin{tabular}{l|c|cccccc|c|c}
    \toprule
    \multirow{2}{*}{Model} & \multirow{2}{*}{Aug.} & \multicolumn{6}{c|}{Minority classes} & \multirow{2}{*}{mAP} & \multirow{2}{*}{NDS}\\
    \cmidrule{3-8}
     & &  Bus & C.V & Trailer & Truck & M.C & \multicolumn{1}{c|}{B.C} & \\
    \midrule 
    \midrule 
    \multirow{3}{*}{CP-Voxel$^\ddagger$~\cite{DBLP:conf/cvpr/YinZK21}} & GT-Aug & 64.4 & \textbf{31.0} & 60.0 & 47.2 & 65.7 & \multicolumn{1}{c|}{41.0} & 63.8 & 68.7 \\
      & Real-Aug$^\dagger$ & 64.5 & 29.0 & 60.1 & 57.3 & 72.2 & \multicolumn{1}{c|}{47.1} & 65.8 & 71.3 \\
      & PGT-Aug & \textbf{68.1} & 29.0 & \textbf{61.7} & \textbf{57.7} & \textbf{74.0} & \multicolumn{1}{c|}{\textbf{48.6}} & \textbf{67.1} & \textbf{72.3} \\
    \midrule
    \multirow{3}{*}{Transfusion-L~\cite{DBLP:conf/cvpr/BaiHZHCFT22}} & GT-Aug & 63.7 & 29.0 & 58.7 & 46.3 & 67.1 & \multicolumn{1}{c|}{\textbf{44.2}} & 63.9 & 68.6 \\
      & Real-Aug$^\dagger$ & 64.3 & \textbf{31.0} & 60.0 & 47.3 & 65.7 & \multicolumn{1}{c|}{41.0} & 63.8 & 68.7 \\
      & PGT-Aug & \textbf{67.3} & 30.1 & \textbf{60.2} & \textbf{56.9} & \textbf{68.2} & \multicolumn{1}{c|}{40.6} & \textbf{65.1} & \textbf{69.9} \\
    \bottomrule
    \end{tabular}
\end{table}
}
\myparagraph{NuScenes \textit{test} set.}
Lastly, we evaluate our model on the nuScenes test set based on CP-Voxel~\cite{DBLP:conf/cvpr/YinZK21} and Transfusion-L~\cite{DBLP:conf/cvpr/BaiHZHCFT22} as a baseline detection model.
We observe in \cref{tab:test} that our model consistently achieves the best NDS and mAP scores (compared to Real-Aug~\cite{DBLP:journals/corr/abs-2305-12853}). Notably, our method shows significant performance gains for minor classes such as Trailer, Truck, Motorcycle, etc. Note that we apply the test time augmentation (TTA) technique for CP-Voxel following their submission to the nuScenes leaderboard. We provide more details about TTA in the supplemental material.

\section{Ablation Studies}
\myparagraph{Quality of Pseudo Labels.}
To evaluate the quality of our generated objects, we measure FID scores by a SE(3)-transformer~\cite{DBLP:conf/nips/FuchsW0W20} trained on the nuScenes dataset. As shown in \cref{tab:fid}, our generated pseudo-LiDAR objects show similar or lower FID scores (than true samples), which may confirm their plausibility and high quality (compare 5th and 6th vs. 7th and 8th columns). We also observe that the pseudo-LiDAR object quality improves with (i) similar Azimuth resolution to the target dataset sensor, (ii) the use of RGB values additionally as input, and (iii) the regularization by group intensity loss. Finally, we used objects generated by Plenoxel with the parameters that gave the best FID score for all 3D object detection tasks. We provide more details in the supplemental material.
\setlength{\tabcolsep}{4pt}
\renewcommand{\arraystretch}{1}
\begin{table}[t]
    \centering
    \caption{\textbf{Quality of Pseudo LiDAR Point Clouds.} FID scores (squared Wasserstein distance between given samples and nuScenes samples, thus lower is better) comparison between variants of our models and public LiDAR datasets, Lyft~\cite{DBLP:conf/corl/HoustonZBYCJOIO20} and A2D2~\cite{DBLP:journals/corr/abs-2004-06320}. Abbr. G.S: Gaussian Splatting} \label{tab:fid}
    \vspace{-1em}
    \resizebox{\linewidth}{!}{
    \begin{tabular}{l|rrrrr|r|rr}
    \toprule 
    & \multicolumn{6}{c|}{Pseudo-LiDAR Point Clouds} & Lyft~\cite{DBLP:conf/corl/HoustonZBYCJOIO20} & A2D2~\cite{DBLP:journals/corr/abs-2004-06320} \\
    \midrule
    Volumetric 3D type & \multicolumn{5}{c|}{Plenoxel~\cite{DBLP:conf/cvpr/Fridovich-KeilY22}} & G.S~\cite{DBLP:journals/tog/KerblKLD23} & - & - \\
    \hline
    Azimuth Resolution (px) & $3600$ & $1080$ &$1080$ & $1080$ & $1080$ & $1080$ & - & - \\
    RGB features & \cmark & \xmark & \xmark & \cmark & \cmark & \cmark & - & - \\
    Group intensity Loss & \cmark & \xmark & \cmark & \xmark & \cmark & \cmark & - & - \\
    \midrule 
    \midrule 
    Bus & 17.7 & 14.6 & 13.1 & \textbf{13.0} & 13.2 & \textbf{11.2} & 8.7 & 19.8  \\
    Construction Vehicle & \textbf{7.0} & 7.5 & 7.6 & 7.5 & 7.6 & \textbf{7.6} & - & 6.0 \\
    Trailer & 20.6 & 12.7 & 11.9 & \textbf{11.9} & 12.2 & \textbf{13.7} & -  & 36.5 \\
    Truck & 8.9 & 8.4 & 7.6 & 7.6 & \textbf{7.3} & \textbf{6.9} & 6.6 & 13.4  \\
    Motorcycle & 20.7 & 2.5 & 7.0 & 7.2 & \textbf{3.7} & \textbf{1.3} & 3.0  & 10.1 \\
    Bicycle & 9.0 & 3.3 & 2.2 & 2.4 & \textbf{2.1} & \textbf{1.8} & 1.8 & 0.7 \\
    \midrule
    Avg. FID Score ($\downarrow$) & 14.2 & 8.2 & 8.3 & 8.3 & \textbf{7.7} & \textbf{7.1} & 4.8 & 14.4 \\
    \hline
    mAP ($\uparrow$) & 63.40 & 63.44 & 63.48 & 63.41 & \textbf{63.52} & \textbf{63.77} &  63.45 & 63.17 \\ 
    NDS ($\uparrow$) & 68.83 & 68.99 & 69.02 & 68.87 & \textbf{69.11} & \textbf{69.35}  & 68.88 & 68.73\\
    \bottomrule
    \end{tabular}}
\end{table}

\myparagraph{Performance based on FID scores.}
We conducted experiments to assess how the quality of generated objects as FID scores affects detection performance. For a fair evaluation, we generated objects in all comparison groups to have the same shape distribution. In conclusion, we observe a clear positive correlation between the average FID scores and detection accuracies, as shown in the last three rows of columns 1 to 5 of \cref{tab:fid}.

\myparagraph{Samples from Other Datasets.} 
For fair comparison, the evaluation was conducted not only though self-validation but also using other public datasets such as Lyft~\cite{DBLP:conf/corl/HoustonZBYCJOIO20} and A2D2~\cite{DBLP:journals/corr/abs-2004-06320}. Despite differences in class ontology and sensor settings from nuScenes, we can prove the effectiveness of our augmentation and confirm a correlation between the FID score and the number of samples in performance gain as shown in \cref{tab:fid}. See the supplementary for matching classes between nuScenes and other datasets and detailed settings.

\myparagraph{Samples from Different Renderer.}
To demonstrate the baseline capability of the proposed pipeline, we applied the results generated by replacing the rendering model with Gaussian-splatting~\cite{DBLP:journals/tog/KerblKLD23} instead of Plenoxels~\cite{DBLP:conf/cvpr/Fridovich-KeilY22}. Due to Gaussian-splatting~\cite{DBLP:journals/tog/KerblKLD23} generates higher-quality objects, so the detection performance improves accordingly as shown in \cref{tab:fid}.

\begin{figure}[t]
  \begin{minipage}[t]{.49\linewidth}
    \centering
    \setlength{\tabcolsep}{4pt}
    \renewcommand{\arraystretch}{1}
    \captionof{table}{\textbf{Mixing ratio between GT and PGT objects}}\label{tab:mixing}
    \resizebox{\linewidth}{!}{
    \begin{tabular}{c|ccccc}%
    \toprule 
    GT:PGT & 0:1 & 1:3 & 1:1 & 3:1 & 1:0 \\
    \midrule \midrule
    mAP ($\uparrow$) & 61.29 & 63.35 & \textbf{63.52} & 63.40 & 63.34 \\ 
    NDS ($\uparrow$) & 67.63 & 69.08 & \textbf{69.11} & 68.78 & 68.71 \\
    \bottomrule
    \end{tabular}}
  \end{minipage}
  \hspace{1em}
  \begin{minipage}[t]{.47\linewidth}
    \centering
    \setlength{\tabcolsep}{4pt}
    \renewcommand{\arraystretch}{1}
    \centering
    \captionof{table}{\textbf{Continuously increasing pseudo-LiDAR data}}\label{tab:rebuttal_bank}
    \resizebox{\linewidth}{!}{
    \begin{tabular}{c|ccc}
    \toprule 
    Size &  bank 1/4 & bank 1/2 & Full bank \\
    \midrule \midrule
    mAP ($\uparrow$)  & 63.18 & 63.43 & \textbf{63.52} \\ 
    NDS ($\uparrow$)  & 68.72 & 68.94 & \textbf{69.11} \\
    \bottomrule
    \end{tabular}} 
    \end{minipage}
\end{figure}

\myparagraph{Mixing Ratio for Domain Alignments.}
Even when applying various methods to reduce the domain gap between the original and generated pseudo objects, there is still a difference between the two sample groups. Therefore, when Out-of-Distribution data is only used, we can see that the detection performance might drop. To alleviate this discrepancy, we used a strategy that mixes the original and pseudo objects to reduce the gap. We experimentally demonstrate that this strategy can mitigate the domain gap, allowing pseudo objects to be used effectively for detection tasks. It is especially effective when the mixing ratio is 1:1, which can be confirmed through \cref{tab:mixing}.

\myparagraph{Performance based on Bank Size.}
We discuss the importance of pseudo-bank size in the detection task.
To examine the impact of varying object shapes, we adjusted the size of the generated bank by decreasing the number of object shapes per class by 1/2 and 1/4.
As shown in \cref{tab:rebuttal_bank}, we found that as the types of objects increased, the performance improved in the validation set.

\begin{table}[t]
    \centering
    \caption{\textbf{PGT Performance on Lyft \textit{val} set.} \textit{Abbr.} E.V: Emergency Vehicle, O.V: Other Vehicle, M.C: Motorcycle, B.C: Bicycle. Ped: Pedestrian} \label{tab:lyft} 
    \vspace{-1em}
    \resizebox{\linewidth}{!}{
    \begin{tabular}{c|c|ccccccc|c}
    \toprule
    \multirow{4}{*}{Model} & \multirow{3}{*}{Class} & \multicolumn{5}{c|}{Target classes} & \multicolumn{2}{c|}{Other classes} & \multirow{4}{*}{mAP} \\
    \cmidrule{3-9}
     &  & Truck & Bus & O.V & M.C & B.C & \multicolumn{1}{|c}{Car} & \multicolumn{1}{c|}{Ped.} & \\
    \cmidrule{2-9}
    & \# of objects in \textit{val} & 2721 & 1653 & 4920 & 187 & 3347 & \multicolumn{1}{|c}{91529} & 4952 \\
    \midrule \midrule
    \multirow{2}{*}{CP-Voxel~\cite{DBLP:conf/cvpr/YinZK21}} & GT-Aug & 19.15 & 20.48 & 31.91 & \textbf{4.54} & 5.31 & \multicolumn{1}{|c}{\textbf{37.14}} & 6.00 & 13.84 \\
    &  PGT-Aug & \textbf{19.85} & \textbf{21.11} & \textbf{31.99} & 4.39 & \textbf{5.48} & \multicolumn{1}{|c}{37.11} & \textbf{6.12} & \textbf{14.01} \\
    \bottomrule
    \end{tabular}}
\end{table}

\myparagraph{Performance on Other Datasets.}
Lastly, we applied the proposed framework to Lyft~\cite{DBLP:conf/corl/HoustonZBYCJOIO20} and KITTI~\cite{DBLP:journals/ijrr/GeigerLSU13} with different sensor configurations, ranges, and the number of sweeps. 
Both datasets did not provide map information, so we modified our scene composition similar to the ground-only composition. 
As shown in \cref{tab:lyft} and \cref{tab:kitti}, we can see that our framework demonstrates performance improvement across all datasets and is applicable in various environments.
\begin{table}[t]
    \centering
    \caption{\textbf{PGT Performance on KITTI \textit{val} set in terms of AP and mAP.}} \label{tab:kitti} 
    \vspace{-1em}
    \resizebox{\linewidth}{!}{
    \begin{tabular}{c|c|ccc|ccc|ccc|c}
    \toprule
    \multirow{5}{*}{Model} & \multirow{3}{*}{Class} & \multicolumn{3}{c|}{Target classes} & \multicolumn{6}{c|}{Other classes} & \multirow{5}{*}{mAP} \\
    \cmidrule{3-11}
     &  & \multicolumn{3}{c|}{Cyclist} & \multicolumn{3}{c|}{Car} & \multicolumn{3}{c|}{Pedestrian} & \\
    \cmidrule{2-11}
    & \# of objects in \textit{val} & 290& 262&56 & 2980 & 5082 & 3116 & 1139 & 605 & 434 &  \\
    \cmidrule{2-11}
    & Difficulty & Easy & Mod. & Hard & Easy & Mod. & Hard & Easy & Mod. & Hard \\
    \midrule \midrule
    \multirow{2}{*}{SECOND~\cite{DBLP:journals/sensors/YanML18}} & GT-Aug & 62.3 & 55.3 & 49.8 & \textbf{91.4} & \textbf{82.4} & \textbf{79.6} & \multicolumn{1}{|c}{87.1} & 67.9 & 63.8  & 68.5\\
      &  PGT-Aug & \textbf{63.3} & \textbf{56.2} & \textbf{50.2} & 90.7 & 82.1 & 79.3 & \multicolumn{1}{|c}{\textbf{90.3}} & \textbf{72.1} & \textbf{67.7} & \textbf{70.1} \\
    \bottomrule
    \end{tabular}}
\end{table}

\myparagraph{Limitations and Future Work.}
We generated objects as close to the real dataset as possible via various methods, but a domain discrepancy still exists between real and pseudo point clouds. Beyond adjusting the mixing ratio, future works will simultaneously consider the domain generalization across various categories and datasets.
\section{Conclusion}
In this paper, we propose PGT-Aug, a low-cost yet effective data augmentation framework for class imbalance in 3D object detection.
To efficiently obtain rare class objects, we start by generating objects from miniatures or web videos at a low cost, then transform them to resemble real LiDAR data, and finally apply map-assistant data augmentation to insert them consistently into the scene information.
We conduct extensive experiments to verify the effectiveness of PGT-Aug and the compatibility for various 3D object detection models and achieve significant improvements on nuScenes, KITTI, and Lyft datasets.

\clearpage
\pagebreak
\begin{center}
    \Large \textbf{Supplementary Material}
\end{center}

%
%
%

\setcounter{section}{0} 
\renewcommand{\thesection}{\Alph{section}}
\section{Dataset Details}

\begin{figure}[ht]
    \begin{center}
    \includegraphics[width=0.9\linewidth]{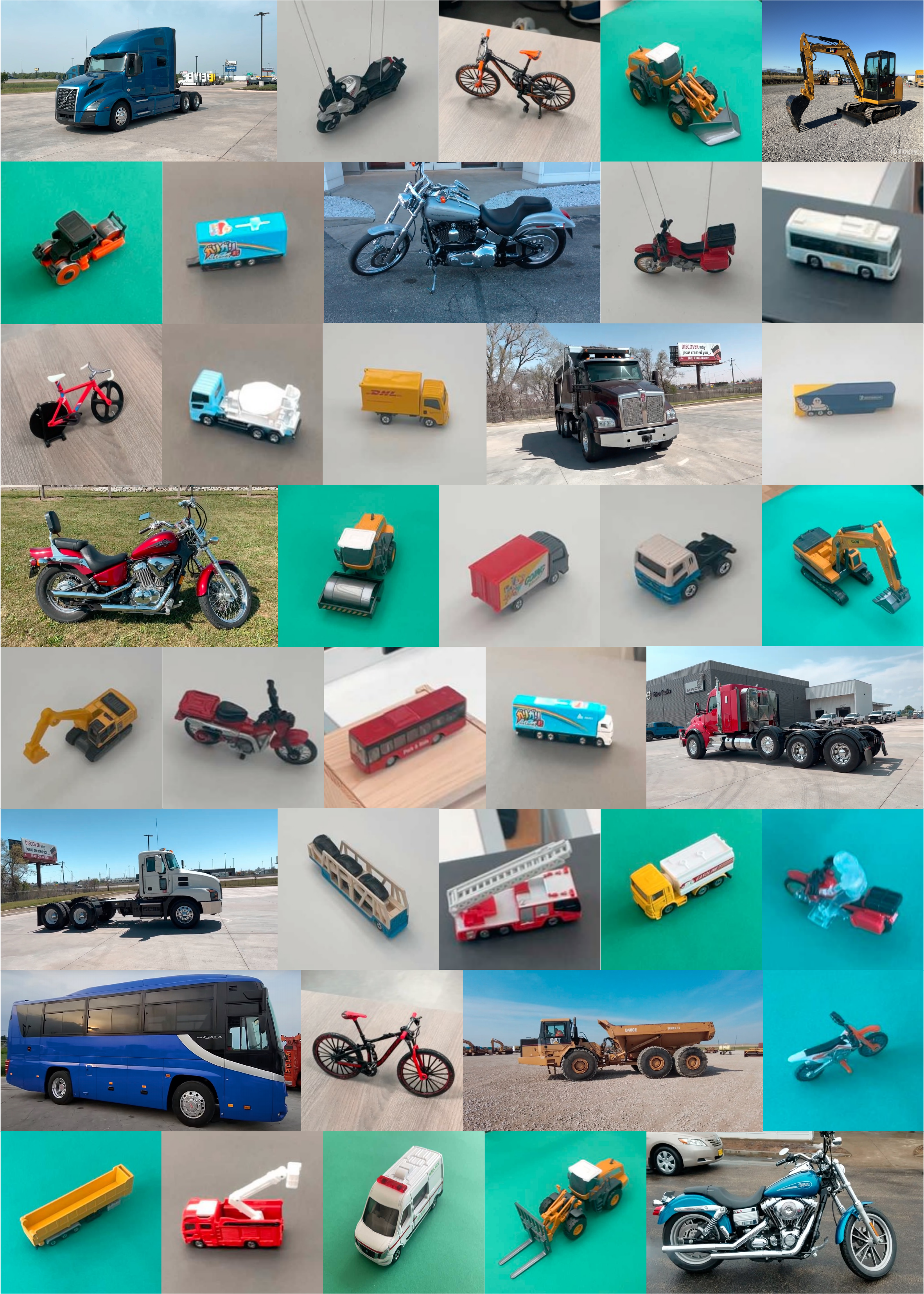}
    \end{center}
    \vspace{-1em}
    \caption{\textbf{Dataset Collection.} We demonstrate our collection of miniature images and crawled web videos. }\label{fig:collection}
\end{figure}

\myparagraph{Data Collections.}
We first provide the number of videos collected from miniatures and web pages as shown in Table~\ref{tab:stat_videos}. We purchased various miniatures within a budget of around \$100, and crawled data using the following keywords on Google: 360$^{\circ}$ camera, surround-view, turntable, sales, and secondhand. Finally, the video dataset consists of 64 trucks\footnote{https://www.youtube.com/@brucknersusedtruckcenterokc}, 21 motorcycles\footnote{https://www.youtube.com/@MHDSuperStore}, and 10 buses\footnote{https://www.youtube.com/@kagamotors} from YouTube, along with additional footage of construction vehicles from Ritchie Bros Auction website\footnote{https://www.rbauction.com/}. We provide examples of our dataset collection in \cref{fig:collection}.

\begin{table}[h]
    \centering
    \caption{Statistics for our data collection by (i) videos capturing surround view of miniatures and (ii) publicly available videos of given minor-class objects. Abbr. C.V: Construction Vehicle, M.C: Motorcycle, B.C: Bicycle}\label{tab:stat_videos}
    \begin{tabular}{lcccccc|c}
    \toprule
        Data Source & C.V & Trailer & Truck & Bus & M.C & B.C & Total\\
    \midrule \midrule
        Miniatures & 10 & 4 & 5 & 3 & 6 & 3 & 31\\
        Public Videos & 2 & - & 64 & 10 & 21 & - & 97 \\
    \bottomrule
    \end{tabular}
\end{table}

\clearpage


\myparagraph{Volumetric 3D Instance Collection.}
To aid in understanding the operation of the proposed framework, we provide all results from input images to output results including foreground segmentation masks, extracted foreground objects, camera poses, and the corresponding dense RGB-colored point clouds.

\begin{figure*}[ht]
    \begin{center}
    \includegraphics[width = 0.875\linewidth]{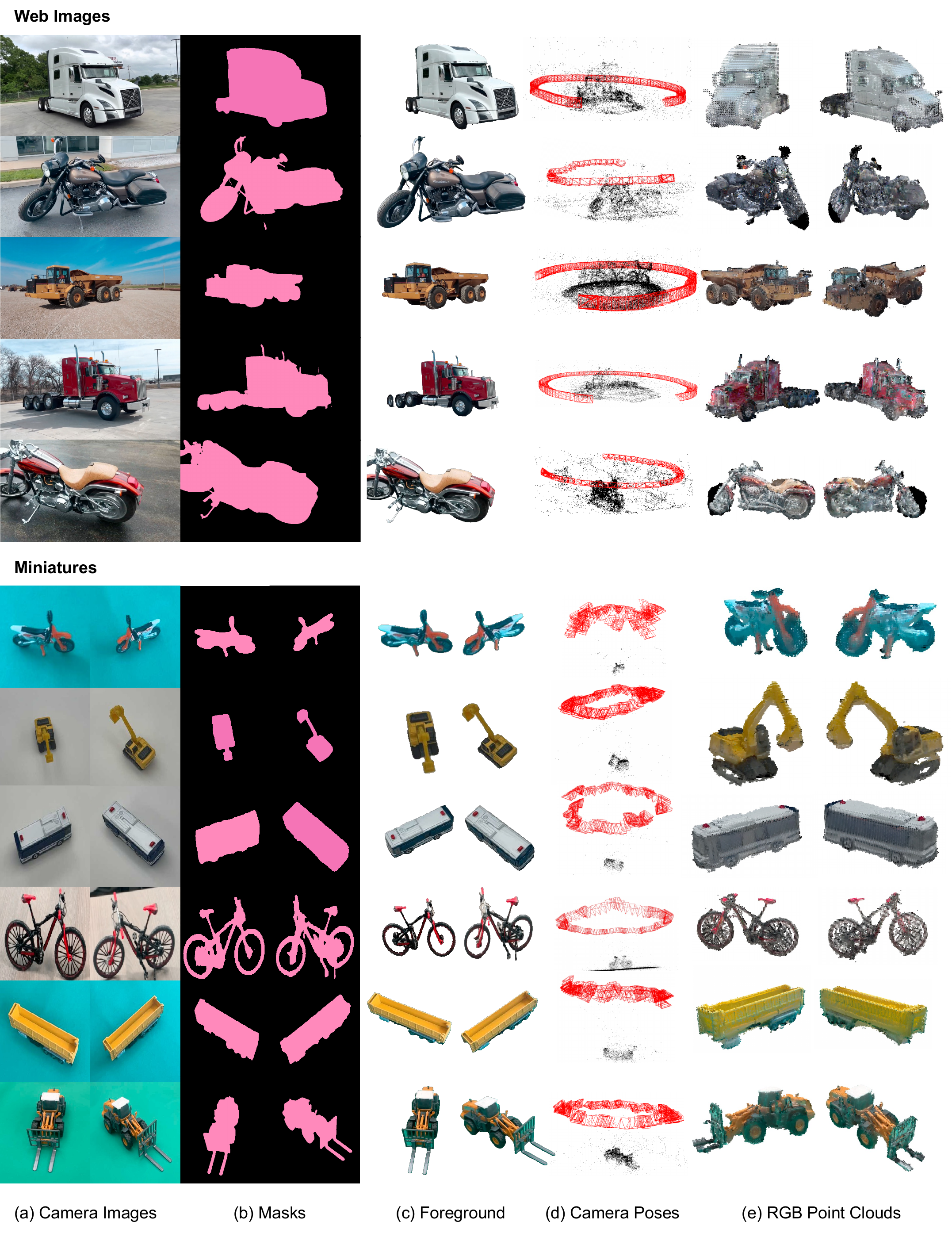}
    \end{center}
    \vspace{-1em}
    \caption{\textbf{Volumetric 3D rendering Results of RGB Point Clouds.} }\label{fig:rgbsamples}
\end{figure*}

\clearpage

\begin{figure}[ht]
    \begin{center}
    \includegraphics[width=0.65\linewidth]{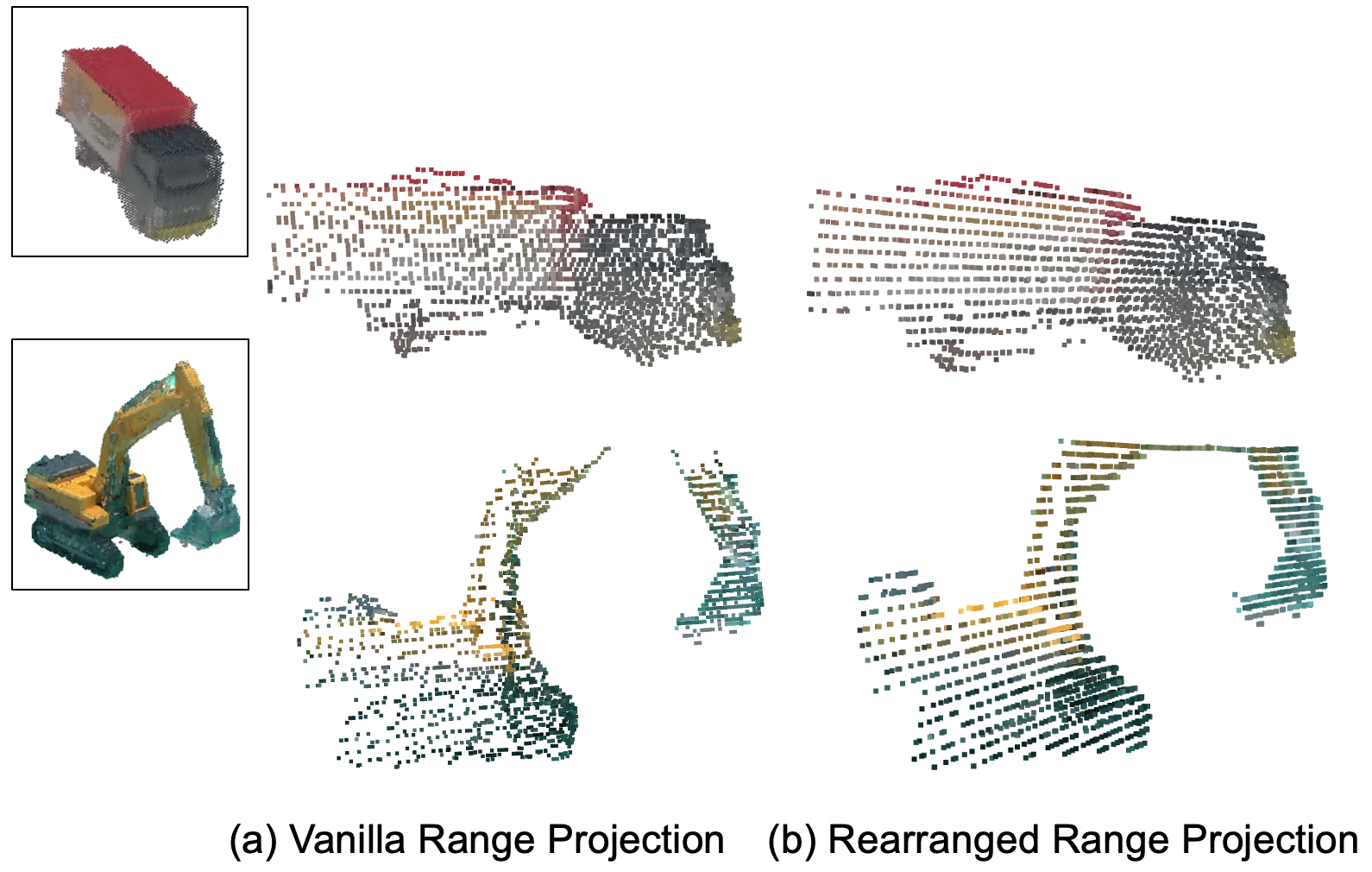}
    \end{center}
    \caption{\textbf{Examples of Generated Point Clouds (a) with and (b) without rearranged range projection.} We use Azimuth resolution: 1080, Vertical FOV: -30$^{\circ}$~10$^{\circ}$, Channels: 32 for sensor configuration.}
    \label{fig:range_proj_supp}
    \vspace{-1em}
\end{figure}

\myparagraph{Object-level Domain Alignment.}
We demonstrate the effect of our rearranged range projection as shown in~\cref{fig:range_proj_supp}. Not only qualitative differences, but also by reflecting the sensor configuration, we can reduce the discrepancy to the actual data domain.

\myparagraph{Generalization by Different Renderers.}
To prove the capability of the proposed framework, we show the generated RGB-colored point clouds and the simulated LiDAR point clouds of yellow loader by using different 2D-3D renderers: Plenoxels~\cite{DBLP:conf/cvpr/Fridovich-KeilY22}, Gaussian-Splatting~\cite{DBLP:journals/tog/KerblKLD23}, and DUSt3R~\cite{DBLP:journals/corr/abs-2312-14132}. Even when applying different models, we can observe that our proposed framework can consistently generate a high quality LiDAR objects.
\begin{figure}[ht]
    \begin{center}
    \includegraphics[width=0.85\linewidth]{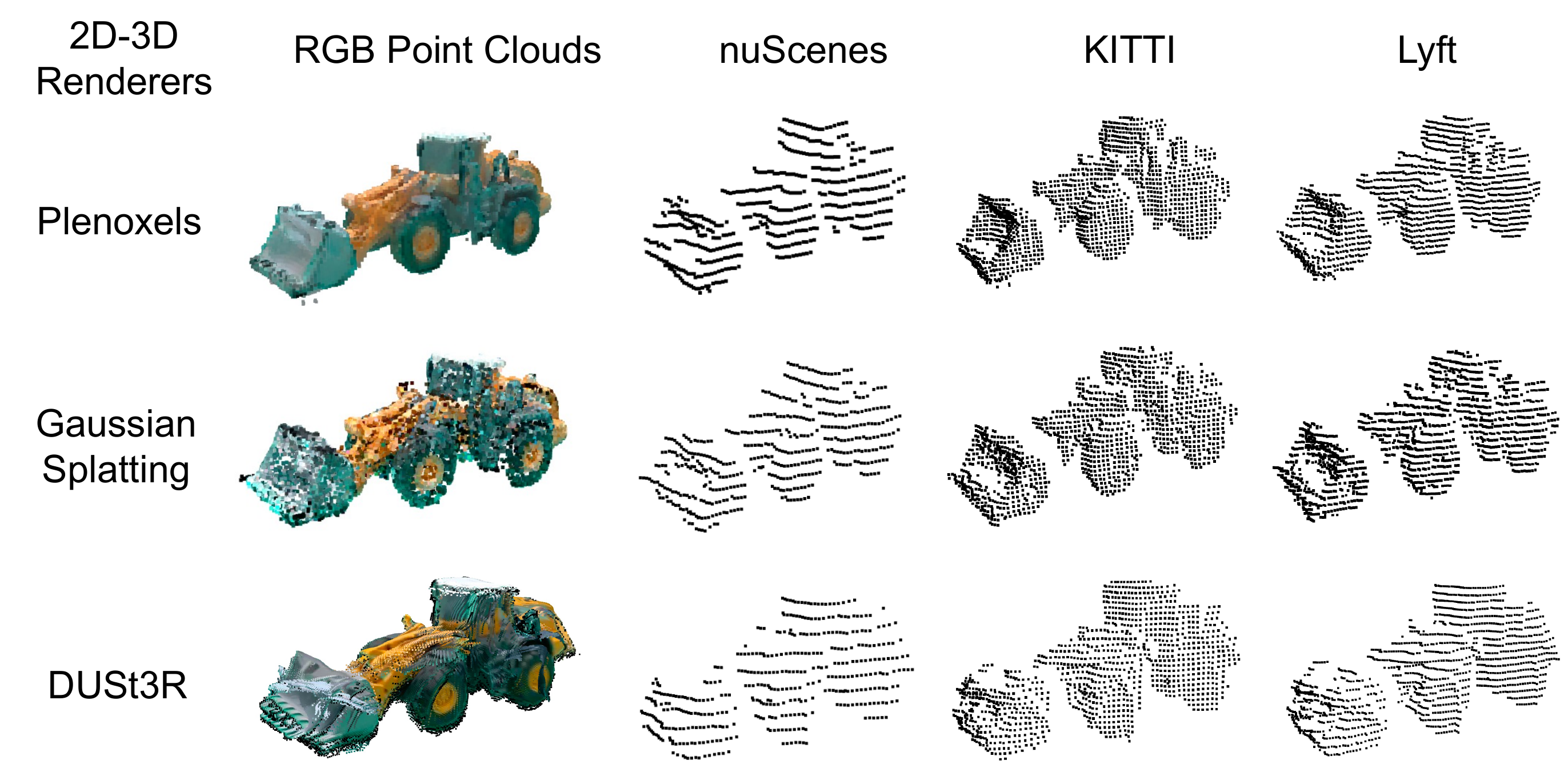}
    \end{center}
    \vspace{-1em}
    \caption{\textbf{Examples of RGB Point Clouds and Generated Pseudo LiDAR from Various 2D to 3D Renderers.}}
    \label{fig:renderes}
    \vspace{-1em}
\end{figure}



\section{Implementation Details}
\myparagraph{View-agnostic Color Representation.}
We provide a pseudo-code for ad-hoc module to obtain representative colors for each voxel grid or point based on the estimated mean color values from different views.
\begin{algorithm}[ht]
  \caption{Pseudocode for estimating mean RGB value of point cloud in a pretrained voxel grid or point cloud.}\label{alg:mean_rgb}
   \hspace*{\algorithmicindent} \textbf{Input} $(H \times W)$ rays from $N$ views, Spherical Harmonic (SH) coefficients $sh$, step size $t_{step}$, Spatial indices of the 3D data structure (Voxels or 3D Gaussians) $l$\\ 
    \hspace*{\algorithmicindent} \textbf{Output} RGB PointCloud
  \begin{algorithmic}[1]
    \For {each $i$-th view in $N$}
        \For{each $j$-th ray in $H \times W$}
            \State Define $RGB_{sum}, count$ to store RGB value and counts at $i$-th view.
            
            \State Calculate ray origin and direction $r_d$, $r_o$ $\leftarrow$ ray $r_{ij}$
            \State Calculate SH basis $sh_{basis}$ with $r_d/\hat{r_d}$.

            \State Calculate the ray $r_{ij}$'s far bound $t_{max}$, and the near bound $t_{min}$.

            \State Set $t$ as $t_{min}$.
            
            \While{$t < t_{max}$}
                \State Extract the position in the range $[t, t + t_{step})$ of ray $r_{ij}$.
                
                \State Obtain spatial index $l_{idx}$ of the position.
               
                \State Obtain SH coefficients $sh_{c} \leftarrow sh[l_{idx}]$.
                
                \State Calculate RGB of voxels or Gaussians by $\Sigma(sh_{c}\times sh_{basis})$.
            
                \State Add RGB value and one to $RGB_{sum}, count$ at index $l_{idx}$, respectively.
            
                \State  Increment $t$ by $t_{step}$
            \EndWhile
        \EndFor
            
        \State Append RGB sum, and counts of voxels or Gaussians in $i$-th view to arrays
    \EndFor
\end{algorithmic}
\end{algorithm}

\myparagraph{Front-back classification.}
For the front-and-back binary classifier, we build a conventional PointNet-based model and train the model in a total of 100 epochs with binary cross-entropy loss. We utilize nuScenes~\cite{DBLP:conf/cvpr/CaesarBLVLXKPBB20} point clouds as the training set and flip the point cloud with a probability of 0.5 to model input. The model transfers input data into 128-dim feature vectors, mapped into scalar probability to classify whether the vehicle heading is towards front or back.

\myparagraph{Intensity Estimator.}
We train the intensity estimation model for each target class.
Our model is based on CycleGAN~\cite{DBLP:conf/iccv/ZhuPIE17}, including class-aware PointNext~\cite{DBLP:conf/nips/QianLPMHEG22} generators and discriminators.
For batch training, we first convert LiDAR objects with more than $256$ points in nuScenes dataset~\cite{DBLP:conf/cvpr/CaesarBLVLXKPBB20} and dense RGB point cloud generated from the volume radiance field into samples with $256$ points by using farthest point sampling.
Then, we randomly chose $300$ sample pairs from RGB and intensity domains and trained the network for $200$ epochs. For two generators $G_{\mathcal{D}_{rgb}\to\mathcal{D}_{int}}$, $G_{\mathcal{D}_{int}\to\mathcal{D}_{rgb}}$, and two discriminators $D_{\mathcal{D}_{int}}$, $D_{\mathcal{D}_{rgb}}$, we use a weighted loss that consists of group intensity loss and cyclic loss,
\begin{equation}
\begin{split}
\mathcal{L}\left(G_{\mathcal{D}_{rgb}\to\mathcal{D}_{int}}, G_{\mathcal{D}_{int}\to\mathcal{D}_{rgb}}, D_{\mathcal{D}_{rgb}}, D_{\mathcal{D}_{int}}\right) = \\     
\mathcal{L}_{\mathrm{GAN}}\left(G_{\mathcal{D}_{rgb}\to\mathcal{D}_{int}}, D_{\mathcal{D}_{int}}, \mathcal{D}_{int}, \mathcal{D}_{rgb}\right) \\ 
+
\mathcal{L}_{\mathrm{GAN}}\left(G_{\mathcal{D}_{int}\to\mathcal{D}_{rgb}}, D_{\mathcal{D}_{rgb}}, \mathcal{D}_{rgb}, \mathcal{D}_{int} \right) \\
+ \mathcal{L}_{\mathrm{cyc}}\left(G_{\mathcal{D}_{rgb}\to\mathcal{D}_{int}}, G_{\mathcal{D}_{int}\to\mathcal{D}_{rgb}} \right) \\
+ \mathcal{L}_{\mathrm{group}}\left(G_{\mathcal{D}_{rgb}\to\mathcal{D}_{int}}, \mathcal{D}_{int}, G_{\mathcal{D}_{int}\to\mathcal{D}_{rgb}}, \mathcal{D}_{rgb}  \right)
\end{split}
\end{equation}

\section{Experiments Details}
\myparagraph{Pseudo Object Bank Details.} To verify the impact of the proposed pipeline on the 3D object detection task on nuScenes dataset~\cite{DBLP:conf/cvpr/CaesarBLVLXKPBB20}, we created a pseudo object bank from RGB points generated by Plenoxel~\cite{DBLP:conf/cvpr/Fridovich-KeilY22}.
We used the intensity estimator trained from the point cloud with RGB features, with Group intensity loss and CycleGAN loss, and performed point filtering and rearrangement according to the sensor configuration of the nuScenes dataset~\cite{DBLP:conf/cvpr/CaesarBLVLXKPBB20}. This pseudo object bank has an FID score of 7.7 on the nuScenes GT object bank (see Table 4 in the main paper).

\myparagraph{Detection Models.}
We provide all configuration YAML files of OpenPCDet~\cite{openpcdet2020} models used in the paper.
Note that we only modified the augmentation part of the configuration for a fair comparison. 

\section{Ablation Studies Details}
To demonstrate the effect of the quality and quantity of the pseudo object banks, we conducted various ablation studies.
All experiments for the 3D object detection task are conducted based on CP-Voxel~\cite{DBLP:conf/cvpr/YinZK21} operating on the input voxel size [0.075, 0.075, 0.2], while maintaining the setting of the remaining hyperparameters from OpenPCDet~\cite{openpcdet2020}.

\myparagraph{Quality of Pseudo Labels.} 
We conducted experiments to assess how plausible and realistic our generated object points are compared to existing real-world datasets. 
To measure FID scores, we use a SE(3)-transformer~\cite{DBLP:conf/nips/FuchsW0W20} model trained on nuScenes dataset.
We first sort the objects in nuScenes in descending order based on the number of points, divide them into 32 groups, and sequentially sample objects with a large number of points, resulting in class sets with 593 objects.
Note that A2D2~\cite{DBLP:journals/corr/abs-2004-06320}, Lyft~\cite{DBLP:conf/corl/HoustonZBYCJOIO20}, and our pseudo GT bank use all available samples. The class mapping between different detection datasets is shown in \cref{tab:classmapping}.
In feature extraction, the object is sampled with 64 points, followed by local grouping with the nearest 32 points.
These features are input into the SE(3)-transformer~\cite{DBLP:conf/nips/FuchsW0W20} to extract the object-aware feature.
\begin{table}[ht]
\centering
\caption{Consistent categories agreement among multiple datasets for FID evaluation and 3D object detection.}
\vspace{-1em}
\begin{tabular}{l|l|l}
\toprule
\textbf{A2D2 Dataset~\cite{DBLP:journals/corr/abs-2004-06320}} & \textbf{Lyft Dataset~\cite{DBLP:conf/corl/HoustonZBYCJOIO20}} & \textbf{nuScenes Dataset~\cite{DBLP:conf/cvpr/CaesarBLVLXKPBB20}}  \\
\midrule
\midrule
Truck&Truck&Truck \\ 
\midrule
Bus&Bus&Bus \\ 
\midrule
MotoBiker, Motorcycle&Motorcycle&Motorcycle \\ 
\midrule
Bicycle, Cyclist&Bicycle&Bicycle  \\ 
\midrule
Trailer&-&Trailer \\ 
\midrule
Utility Vehicle & - & Construction Vehicle \\
\midrule
 & Animal & \\
Caravan Transporter & Emergency Vehicle & - \\
& Other Vehicle & \\
\bottomrule
\end{tabular}
\label{tab:classmapping}
\end{table}

\myparagraph{Samples from Other Datasets.} Besides pseudo LiDAR generation for nuScenes dataset, we also generated object banks for KITTI~\cite{DBLP:journals/corr/abs-2004-06320} and Lyft~\cite{DBLP:conf/corl/HoustonZBYCJOIO20} to demonstrate PGT-Aug's effectiveness on different detecction benchmarks. We adopt perception range of $x$ from -40 to 40m and $y$ from 0 to 70m at the interval of 5m for KITTI~\cite{DBLP:journals/corr/abs-2004-06320}, and adopt range from -80m to 80m for Lyft~\cite{DBLP:conf/corl/HoustonZBYCJOIO20} according to dataset specific evaluation schemes. We use 13 heading directions (from -180$^{\circ}$ to 180$^{\circ}$ at intervals of 30$^{\circ}$). We augment pseudo LiDAR for [cyclist] class for KITTI~\cite{DBLP:journals/corr/abs-2004-06320}, and augment [truck, motorcycle, bicycle, bus, other vehicle] classes for Lyft~\cite{DBLP:conf/corl/HoustonZBYCJOIO20}.   

\clearpage

\section{Qualitative Analysis for 3D Detection Task}
\begin{figure*}[!ht]
    \begin{center}
    \includegraphics[width = 0.9\linewidth]{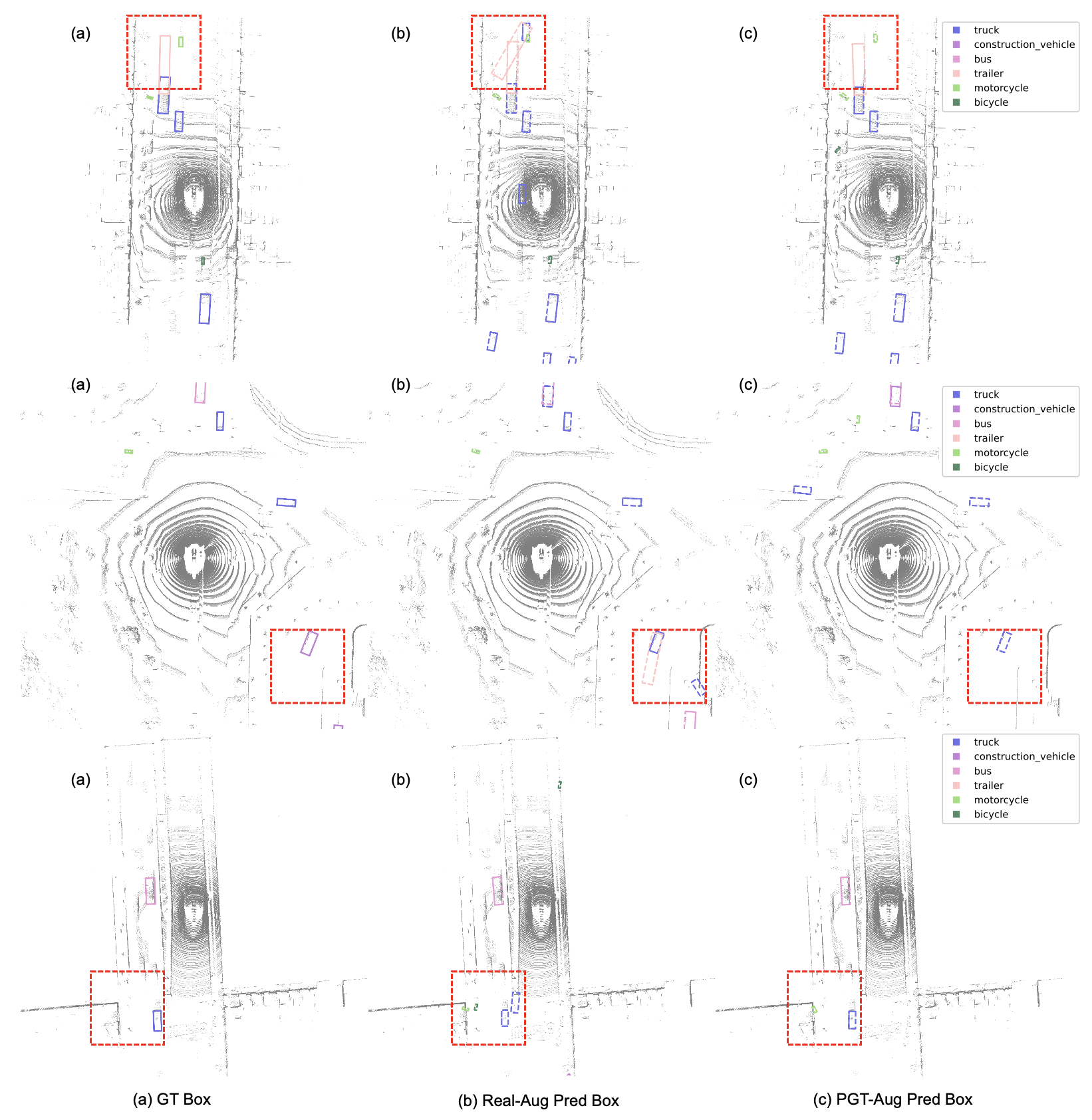}
    \end{center}
    \vspace{-1em}
    \caption{\textbf{Detection output comparison between Real-Aug and PGT-Aug (ours). We provide bounding boxes of (a) ground truth and detected by (b) Real-Aug and (c) PGT-Aug.} }\label{fig:vizdet}
    \vspace{-1em}
\end{figure*}

We provide additional qualitative results of the model trained with the proposed PGT-Aug in \cref{fig:vizdet}. For a fair comparison, we provide ground truths boxes for target classes~(see (a)) and results from models with Real-Aug~\cite{DBLP:journals/corr/abs-2305-12853} and PGT-Aug (ours), which leverages pseudo LiDAR samples~(compare (b) vs. (c)).

\clearpage

\section{Discussion Towards Better Performance}
We introduce some novel techniques for improving the effectiveness of PGT-Aug that were not applied in main paper for a fair comparison. \textit{We expect that these methods will be helpful for future researches}.

\myparagraph{Ray tracing and Distance filtering.} 
Ray tracing can simulate a partially occluded form of objects when obstacles exist between the newly inserted object and the LiDAR sensor location by a range projection. Through this method, we can generate the realistic LiDAR point cloud scene.
For distance filtering, we observed that ground truths located outside the perception range can increase the class uncertainty due to sparsity. Therefore, we can reduce the sample uncertainty by eliminating those samples in the ground truth bank. In this experiment, we set a distance threshold of $54m$.

\myparagraph{Bandit Algorithm for Object Placement.}
The advantage of our pipeline is that it can generate fully volumetric objects, allowing insertion at any location.
To maximize the advantage, we use Thompson Sampling~\cite{DBLP:journals/ftml/RussoRKOW18} to determine the candidate positions for pseudo LiDAR samples during the training of detectors.
We formulate this position decision as a bandit problem of selecting cells of a grid representing a quantized space around the ego vehicle.
First, for cells with the ground truth box, if the confidence score of the prediction is greater than 0.5, the detector is considered to have performed well in the cell area, and the success count of the cell increases.
On the other hand, if the confidence score of the negative samples is greater than 0.5, it is considered a failure, and the failure count of the cell increases. For each cell, Thompson Sampling is performed by setting the alpha of the beta distribution as the fail count and the beta as the success count.
After that, the top $n$ cells with the highest probability are selected.
This process allows us to insert objects primarily where the detector fails during training.
We set the grid size $0.075 \times 8 m$ in this experiment.
\setlength{\tabcolsep}{4pt}
\renewcommand{\arraystretch}{1.3}
\begin{table}[h]
    \centering
    \caption{\textbf{Detection performance comparison for additional modules on nuScenes \textit{val} set in terms of AP, mAP, and NDS}. We use CP-Voxel~\cite{DBLP:conf/cvpr/YinZK21} and Transfusion-L~\cite{DBLP:conf/cvpr/BaiHZHCFT22} as a baseline model.}   \label{tab:supp}
    \vspace{-1em}
    \small
    \resizebox{0.95\linewidth}{!}{
    \begin{tabular}{lcccc|cccc}
    \toprule
    \multirow{2}{*}{Model} & \multicolumn{4}{c|}{CP-Voxel~\cite{DBLP:conf/cvpr/YinZK21}} & \multicolumn{4}{c}{Transfusion-L~\cite{DBLP:conf/cvpr/BaiHZHCFT22}} \\\cmidrule{2-9}
     & A & B & C & D & E & F & G & H\\\midrule
    w/ PGT-Aug & \cmark & \cmark & \cmark & \cmark & \cmark & \cmark & \cmark & \cmark\\
    w/ Ray Tracing & \xmark & \cmark & \cmark & \cmark & \xmark & \cmark & \cmark & \cmark \\
    w/ Distance Filtering & \xmark & \xmark & \cmark & \cmark & \xmark & \xmark & \cmark & \cmark \\
    w/ Thompson Sampling & \xmark & \xmark & \xmark & \cmark & \xmark & \xmark & \xmark & \cmark \\
    \midrule\midrule
    mAP ($\uparrow$, \textit{for all 10 classes}) & 63.5 & 63.8 & 63.6 & 63.7 & 64.2 & 65.0 & 64.4 & 65.2\\
    NDS ($\uparrow$, \textit{for all 10 classes}) & 69.1 & 69.3 & 69.0 & 69.3 & 68.8 & 69.2 & 68.7 & 69.4\\ 
    \bottomrule
  \end{tabular}}
\end{table}

%
%
\bibliographystyle{splncs04}
\bibliography{main}

\end{document}